\documentclass[10pt,twocolumn,letterpaper]{article}

\usepackage{wacv}              

\usepackage{graphicx}
\usepackage{amsmath}
\usepackage{amssymb}
\usepackage{booktabs}
\usepackage{multirow}

\usepackage[pagebackref,breaklinks,colorlinks]{hyperref}

\usepackage[capitalize]{cleveref}
\crefname{section}{Sec.}{Secs.}
\Crefname{section}{Section}{Sections}
\Crefname{table}{Table}{Tables}
\crefname{table}{Tab.}{Tabs.}

\def\wacvPaperID{1200} 
\def\confName{WACV}
\def\confYear{2024}

\begin{document}

\title{SpectralCLIP: Preventing Artifacts in Text-Guided Style Transfer\\ from a Spectral Perspective}

\author{Zipeng Xu\textsuperscript{1}\footnotemark[1] \quad Songlong Xing\textsuperscript{1}\footnotemark[1] \quad
Enver Sangineto\textsuperscript{2} \quad
Nicu Sebe\textsuperscript{1} \\
\textsuperscript{1}University of Trento, Italy \quad  \textsuperscript{2}University of Modena and Reggio Emilia, Italy \\
{\tt\small \{zipeng.xu, songlong.xing, niculae.sebe\}@unitn.it \quad  enver.sangineto@unimore.it}
}


\twocolumn[{
\maketitle
\begin{center}
    \captionsetup{type=figure}
    \includegraphics[width=.98\linewidth]{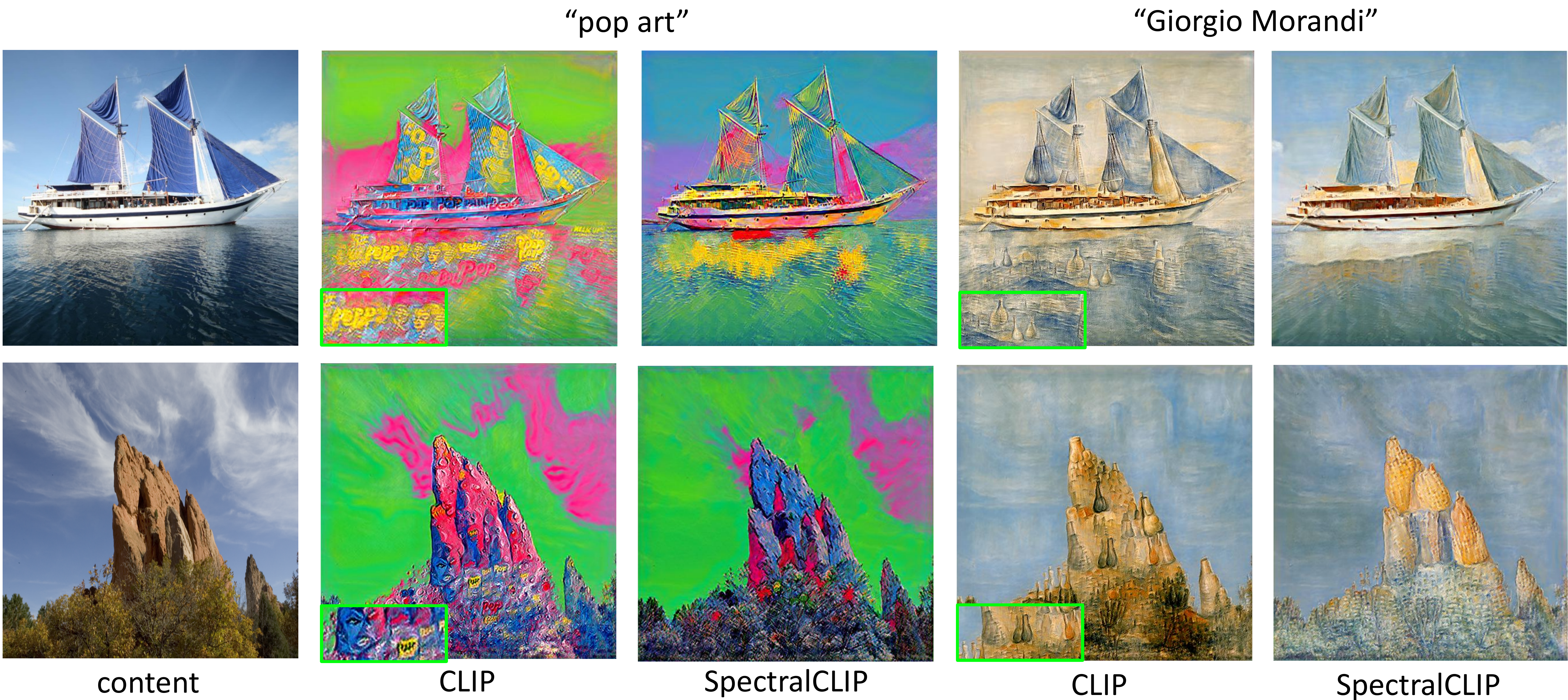}
    \captionof{figure}{Text-guided style transfer results using either a CLIP-based image-text similarity  or  SpectralCLIP. While using the standard CLIP leads to the generation of undesirable artifacts (written words and unrelated visual entities as \textit{\textcolor{green}{highlighted}}), the proposed SpectralCLIP prevents the artifacts while maintaining other style features.
    }
    \label{fig:teaser}
\end{center}
}]

\renewcommand{\thefootnote}{\fnsymbol{footnote}}
\footnotetext[1]{The two authors contributed equally to this paper.}

\begin{abstract}
\vspace{-0.3cm}
Owing to the power of vision-language foundation models, e.g., CLIP, the area of image synthesis has seen recent important advances.
Particularly, for style transfer, CLIP enables transferring more general and abstract styles without collecting the style images in advance, as the style can be efficiently described with natural language, and the result is optimized by minimizing the CLIP similarity between the text description and the stylized image.
However, directly using CLIP to guide style transfer leads to undesirable artifacts (mainly written words and unrelated visual entities) spread over the image. 
In this paper, we propose SpectralCLIP, which is based on a spectral representation of the CLIP embedding sequence, where most of the common artifacts occupy specific frequencies. By masking the band including these frequencies, we can condition the generation process to adhere to the  target style properties (\eg, color, texture, paint stroke, etc.) while excluding the generation of larger-scale structures corresponding to the  artifacts.
Experimental results show that SpectralCLIP prevents the generation of artifacts effectively in quantitative and qualitative terms, without impairing the stylisation quality.
We also apply SpectralCLIP to text-conditioned image generation and show that it prevents written words in the generated images. 
Our code is available at \href{https://github.com/zipengxuc/SpectralCLIP}{https://github.com/zipengxuc/SpectralCLIP}.
\vspace{-0.5cm}
\end{abstract}

\section{Introduction}
\label{sec:intro}

Style transfer is about transforming the overall appearance of a given {\em content image} to adhere to a specific style while preserving its content. Starting from the pioneering paper of Gatys \etal. \cite{gatys2016image}, this task has attracted a growing interest in the scientific community because of its large application interest (\eg, in the e-commerce or the entertainment industry, etc.).
While most of the methods proposed so far extract the style information from a {\em reference image} \cite{gatys2016image, huang2017arbitrary, park2019arbitrary,Chen_2021_CVPR, Lin_2021_CVPR, Deng_2022_CVPR, deng2022stytr2, wang2022aesust, chen2021artistic, DBLP:conf/cvpr/LiuSCBZSL0N21, cheng2021style, yao2019attention, liu2021adaattn, ulyanov2017improved, wang2020diversified} or a set of reference images \cite{sanakoyeu2018styleaware, zhu2017unpaired, kotovenko2019content}, 
very recently, with the emergence of vision-language foundation models, a few approaches have  started investigating the use of a {\em textual description} of the target style \cite{gal2022stylegan,kwon2022clipstyler,Xu2023StylerDALLE}. The main idea is to
describe the target style with a natural language sentence (\eg, ``pop art'') which is used to condition the content image  transformation. 
The main advantage of this approach is that natural language sentences can describe more general and more abstract style characteristics that can hardly be extracted from a single reference image. Moreover,  this way, it is possible to indirectly exploit the knowledge contained in the vision-language foundation model, which is usually pre-trained using hundreds of millions of image-text pairs.

Kwon \etal proposed CLIPstyler \cite{kwon2022clipstyler}, which utilises the power of CLIP for text-guided style transfer for arbitrary images, demonstrating a broader range of styles and higher transfer quality than previous work based on reference images. 
However, as pointed out in \cite{kwon2022clipstyler}, this method tends to generate images with over-specific artifacts. To distinguish, we define two types of artifacts: textual and visual artifacts. 
Visual artifacts are over-specific entities drawn on the generated image. In the example of \cref{fig:teaser}, when the style is ``pop art'', CLIPstyler adds red lips and faces onto the image.
Textual artifacts are written words, typically from the textual prompt describing the desired style, that appear on the generated image in an unwanted manner. Examples can also be seen in \cref{fig:teaser}, where the word ``pop'' is spread over the generated image when CLIPstyler transfers the image with the style of ``pop art''. 
The presence of this type of artifact is largely due to the entanglement of visual concepts and written texts inherent in CLIP \cite{Torralba-Disentangling}. This entanglement issue in CLIP has been shown to be problematic and prevalent in a variety of CLIP application scenarios, including zero-shot classification and text-guided generation. The study and alleviation of effects resulting from such undesirable entanglement is a significant direction that has attracted increasing research attention \cite{goh2021multimodal,DBLP:conf/iclr/LemesleSPASO22,Torralba-Disentangling}.

In this paper, we propose to prevent artifact generation in text-guided style transfer using a spectral approach.
Spectral analysis has been used to analyze temporal or spatial variations of signals~\cite{gonzalez2018digital}. 
Recently, researchers in the field of natural language processing (NLP) have proven its effectiveness in capturing linguistic information at different granular levels \cite{Prism,muller-eberstein-etal-2022-spectral}. 
Tamkin \etal \cite{Prism} show that the sequence of textual tokens input to a Transformer network \cite{attention-is-all-you-need} contains structures at different scales: \eg, the word scale, the sentence scale, the document scale, etc. These scales correspond to different frequencies in the changes of the values of the neurons' activations and can be isolated using the coefficients obtained by applying a DCT to the  neuron activation sequence \cite{Prism}.
Intuitively, analogously to a sequence of linguistic tokens containing a hierarchy of semantic levels (\ie from word level up to the document level), the constituent patches of an image also contain information at different levels. 
Similarly to \cite{Prism},
we sort the frequency components into several continuous bands.
After analysing the patterns of artifacts present in CLIPstyler-generated stylised images, we find that these artifacts are highly related to certain frequency bands, and that by masking out these frequency components, we can remove the artifacts effectively without hurting the quality of stylised images. 
Hence, we propose SpectralCLIP to mask out those frequency bands, which implements a spectral filtering layer on top of the last layer of the CLIP vision encoder.
We conduct experiments that verify the following points: 
(i) we experiment with many types of styles and find SpectralCLIP can effectively reduce both visual and textual artifacts while maintaining the target style well;
(ii) we conduct a user study of 30 participants to compare the visual quality in terms of the overall style and the artifact-free performance of generated images, and find that our generated images are preferred by \textbf{\textit{55.28\%}} and \textbf{\textit{74.44\%}} of the users in terms of overall quality and artifact-free performance, respectively (Sec. \ref{sec:forget-to-spell}); 
and (iii)  we also leverage the `learn-to-spell' CLIP (the CLIP subspace that focuses on written texts in the image) \cite{Torralba-Disentangling} to
quantitatively validate that SpectralCLIP efficiently reduces textual artifacts (Sec. \ref{sec:StyleTransferResults}) as the score w.r.t. written texts is notably reduced.
In addition, we employ SpectralCLIP for text-guided image generation (Sec. \ref{image_generation}) and show it effectively prevents written words on the generated images. 

To conclude, the contributions of this paper are:
\begin{itemize}
    \item We propose SpectralCLIP to prevent both textual and visual artifacts in CLIP-guided style transfer. The effectiveness of SpectralCLIP has been verified on multiple styles through qualitative results, quantitative results, and a user study.
    \item SpectralCLIP is the first work to use spectral filtering in vision-language models. Other than solving the artifacts issues in CLIP-guided image style transfer, it also gives a new perspective on the disentangling of written texts and visual concepts in the CLIP space.
    \item To emphasize the generality of  SpectralCLIP, we show that it can reduce the textual artifact generation also when used in a non style-transfer task and jointly with a completely different generator  based on VQGAN.
\end{itemize} 


\section{Related Work}
\label{sec:Related}
\textbf{Reference Image-Based Style Transfer.}
After a few initial works on style transfer~\cite{efros2001image, hertzmann2001image},
Gatys et al. \cite{gatys2016image} propose to use the Gram matrix of a convolutional  network to represent the target style extracted from a single reference image.
Following this paradigm, 
multiple aspects have been successively explored, ranging from \eg, arbitrary style transfer~\cite{huang2017arbitrary, gu2018arbitrary, park2019arbitrary}, diversified style transfer~\cite{ulyanov2017improved, wang2020diversified}, attention mechanisms to fuse  style and content~\cite{yao2019attention, park2019arbitrary,liu2021adaattn}, reducing artifacts~\cite{chen2021artistic, DBLP:conf/cvpr/LiuSCBZSL0N21, cheng2021style, wang2022aesust}, increasing the content persistence~\cite{li2019learning,wu2022ccpl}, and many others.
However, it is tricky to {\em generalize} the description of an abstract  style (\eg, ``pop art'' or ``warm and calm'') from a single  reference image.  
For this reason, a line of work is based on using a (large) {\em set} of reference images of the target style~\cite{zhu2017unpaired,sanakoyeu2018styleaware,kotovenko2019content, chen2021dualast}.
In contrast, humans can usually understand a style using one or a few words, thanks to their knowledge and the relation they learned between words and visual appearance. 
Massively trained vision-language models like CLIP~\cite{radford2021learning} now make it possible to emulate this human process, and text-guided style transfer approaches can avoid collecting a dataset for each target style, with a simple textual query.

\textbf{CLIP-Guided Image Synthesis.}
The CLIP space has been largely used for image synthesis (\eg, image generation~\cite{crowson2022vqgan, pinkney2022clip2latent, schaldenbrand2022styleclipdraw} and image manipulation
~\cite{styleclip, bau2021paint,xu2022predict,choi2022referring}).
However, when using CLIP for a text-guided style transfer task, a challenging aspect  is to preserve the content while changing the style, since the style textual description usually does not contain any reference to this content.
To solve this problem, 
Gal \etal~\cite{gal2022stylegan} propose a directional loss and fine-tune a StyleGAN~\cite{stylegan, stylegan2} pre-trained using images of a specific domain.
CLIPstyler \cite{kwon2022clipstyler} extends  this approach  to an open-domain scenario and uses multiple  patches.
However, the multi-patch directional loss leads to 
the generation of textual artifacts and 
 ``over-specification'', where the latter refers to over-specific visual artifacts which locally remind of the  textual  description of the style
\cite{kwon2022clipstyler} (\cref{sec:intro}).
Most of the experiments shown in this paper are based on a CLIPstyler baseline, and we show that using SpectralCLIP for the directional loss computation, we can largely alleviate  both the visual and textual artifact problem.
Finally, 
Materzynska \etal~\cite{Torralba-Disentangling} 
analyse the  text-image entanglement problem in the CLIP space, and learn 
orthogonal projections 
(``forget-to-spell''  and ``learn-to-spell'')
of this space  to disentangle the two modalities. 
We empirically show that using the ``forget-to-spell'' projection on a CLIPstyler baseline, we can 
indeed reduce the textual artifacts. In contrast, the proposed SpectralCLIP can reduce the generation of both the visual  and the textual artifacts.


\section{Method}
\label{sec:Method}
SpectralCLIP is based on computing a text-image similarity using a frequency filter of the CLIP representations.  In this section, we first describe how this filtering  is obtained (Sec.~\ref{sec:Spectral-representation}), and then we show  how it can be plugged into existing text-based generative approaches (Sec.~\ref{sec:metrics}), and finally how the band filters are selected (Sec.~\ref{sec:band-selection}).

\begin{figure}[t]
  \centering
  \vspace{-0.2cm}
   \includegraphics[width=.9\linewidth]{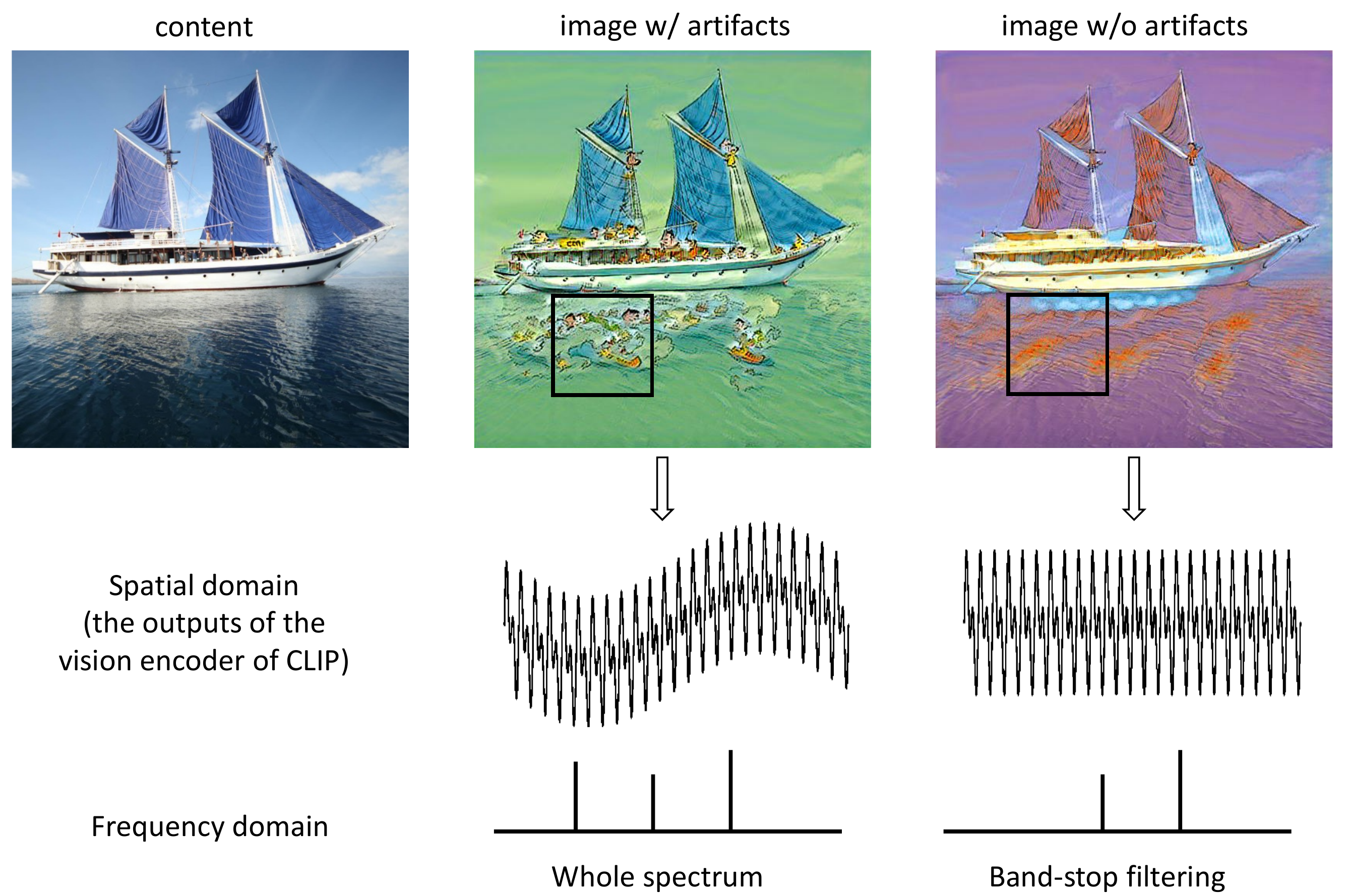}
   \caption{An illustration of SpectralCLIP. To transfer a ``cartoon'' style to the leftmost content image, CLIPstyler generates many cartoon-like artifacts, spreading over the whole image (central figure). The corresponding spectral representation is a composition of frequencies with different periods. Removing the  frequencies corresponding to the artifact scales (SpectralCLIP) prevents the generation of these unwanted artifacts (right figure).}
   \label{fig:method}
   \vspace{-0.2cm}
\end{figure}

\subsection{Spectral based filtering}
\label{sec:Spectral-representation}
Given an image $I$, we use the CLIP vision encoder $E_v(\cdot)$ to represent $I$ with a grid of $k \times k$ vectors which  either are extracted from the last convolutional layer of a ResNet \cite{he2016deep} or correspond to the final embeddings of a Vision Transformer \cite{ViT}. Our method is independent of the specific encoder architecture and can be applied to both types of networks. In the experiments of this paper, following \cite{kwon2022clipstyler}, we used a  ViT-B/32 \cite{ViT}, pre-trained by the authors of CLIP and then frozen. Since a ViT-based encoder also includes a class token, we get $n = 1+k^2$ vectors, which we  flatten into a sequence:  $V = E_v(I)$, where $V = \{ \pmb{v}_0, ..., \pmb{v}_i, ..., \pmb{v}_{n-1} \}$. 
Despite each $\pmb{v}_i \in \mathbb{R}^d$ being a $d$-dimensional vector, following \cite{Prism}, a spectral representation of $V$ can be obtained separately considering each dimension $j$ ($0 \leq j \leq d-1$), which, in our case,
corresponds to the   $j$-th channel of the CLIP embedding. For a given $j$,  if $x_i^{(j)} = \pmb{v}_i[j]$ is the $j$-th component of $\pmb{v}_i$, the corresponding (scalar-valued) sequence  $X^{(j)} = \{ x_0^{(j)}, ..., x_{n-1}^{(j)} \}$ can be represented in the frequency domain using, \eg,  the DCT-II variant of DCT:
\vspace{-0.1cm}
\begin{equation}
    \label{eq:dct}
    f_m^{(j)} = \sum_{i=0}^{n-1} x_i^{(j)} \cos \left[ \frac{\pi m}{n} (i + 1/2) \right],
    \vspace{-0.2cm}
\end{equation}
where $m = 0, ..., n-1$, and $f_m^{(j)}$ is the coefficient of the $m$-th frequency and represents the contribution of the corresponding cosine wave in the ``signal'' discretely  sampled by the sequence $X^{(j)}$. Different frequencies describe different cosine waves, and each frequency  period (i.e., the number of elements of $X^{(j)}$ it takes to complete a full cycle) corresponds to the scale of the change in the activation of the $j$-th neuron.
Tamkin \etal \cite{Prism} observe that, in the natural language processing domain, these scale changes usually correspond to different structures contained in the input document: \eg, the single word structure corresponds to the highest frequencies, while medium-level frequencies correspond to sentences, etc. Analogously, since, in our domain,
artifacts are  relatively large-scale visual structures
appearing repeatedly throughout the image (\cref{sec:intro}),  we separate those frequencies most likely corresponding to the artifacts from the other frequencies containing useful style information (texture, color, etc.). To do so, we use a {\em band-stop} filter (see \cref{fig:method}), 
inspired by  periodic noise removal techniques in spectral-based image processing \cite{gonzalez2018digital}.

Concretely, 
we stack all the frequencies of all the $d$ channels in a single $d \times n$ matrix $F$, where the $j$-th row $F_j$ contains the $n$ DCT coefficients in \cref{eq:dct}. 
Then, for each target style, we define a binary filter $\pmb{b} \in \{0,1\}^n$, which contains zero elements only in specific bands  
(see \cref{sec:band-selection} for details).
We use $\pmb{b}$ to zero out those  columns in $F$ which should be filtered:
\vspace{-0.1cm}
\begin{equation}
    \label{eq:filtering}
    S = F \odot M[\pmb{b}],
\end{equation}
where $\odot$ is the Hadamard product, and $M[\pmb{b}]$ is a $d \times n$  matrix in which all the elements are ones except those corresponding to the columns in $\pmb{b}$, which are zeros.

$S$ is the spectral representation of $V$, in which frequencies $\pmb{b}$ are {\em ignored}. 
Note that, differently from \cite{Prism}, where the spectrum of each neuron is individually filtered, in our case $\pmb{b}$ is 
uniformly used for all the $d$ dimensions. This is because an artifact is a complex visual structure, most likely simultaneously involving different dimensions of the CLIP space.
$S$ is finally back-projected into the original CLIP space using the inverse DCT (IDCT), obtaining 
$\hat{V}= \{ \hat{\pmb{v}}_0, ...,  \hat{\pmb{v}}_{n-1} \}$.
$\hat{V}$ is a representation of $I$ which 
can be used jointly with different metrics (\eg, the Euclidean metric or a cosine similarity, etc.) to compute a CLIP-based similarity between images or between images and text {\em which is not influenced by the frequencies in} $\pmb{b}$. This way, we can condition the generation process using a textual sentence (\cref{sec:metrics}) while simultaneously ignoring those frequencies corresponding to the artifact generation.


\subsection{Computing an image-text similarity}
\label{sec:metrics}

In this section, we show how the proposed spectral-based filtering  of an image representation (\cref{sec:Spectral-representation}) can be used to condition a generative process  
and plugged into existing text-conditioned generative frameworks with negligible  modifications of the original approaches. In our experiments, we  use both CLIPstyler \cite{kwon2022clipstyler} (a state-of-the-art text-guided style transfer method, see \cref{sec:intro,sec:Related}) and  the VQGAN+CLIP method \cite{crowson2022vqgan} adopted in \cite{Torralba-Disentangling}.

{\bf VQGAN+CLIP} \cite{crowson2022vqgan} is a text-to-image generation approach based on 
VQGAN  discrete latent codes \cite{DBLP:conf/cvpr/EsserRO21}. 
The latter  are randomly sampled and
then optimized using  the cosine similarity between the
CLIP embedding of the generated image ($\pmb{z}_v$) and the CLIP embedding
of a textual prompt ($\pmb{z}_t$). 
The only thing we need to change to use SpectralCLIP in this framework is the image representation. To do so, we use 
$\pmb{z}_v = \hat{\pmb{v}}_{0}$, where $\hat{\pmb{v}}_{0}$ is the representation of the class token in $\hat{V}$ (\cref{sec:Spectral-representation}). 
Note that $\hat{\pmb{v}}_{0} \neq \pmb{v}_{0}$ because the frequencies in $\pmb{b}$ have been removed. 
Other possible choices can be, \eg, using an average pooling of $\hat{V}$ or a linear projection of the concatenation of all the elements of $\hat{V}$ into a vector of the same dimensions as $\pmb{z}_t$. Following \cite{crowson2022vqgan} we use the class token which is a simple and effective solution.

{\bf CLIPstyler} \cite{kwon2022clipstyler} is based on a U-Net generator \cite{DBLP:conf/miccai/RonnebergerFB15} which, given a content image $I_c$ as input, generates a style-transferred image $I_s$. To condition the generation process on a style textual description $s$ (where $s$ is a natural language sentence), the embedding of $s$, obtained using the textual CLIP encoder,  is compared with the textual embedding of a fixed sentence (``Photo''). The difference between these two textual embeddings should have the same direction of the difference between the visual embedding of $I_c$ and $I_s$. This {\em directional CLIP loss}, initially proposed in \cite{gal2022stylegan}, is further developed in CLIPstyler by introducing patch-level comparisons. We adopt exactly the same framework, and the only necessary change to use SpectralCLIP in CLIPstyler is to replace the standard CLIP visual embedding of an image (or an image patch) with our filtered representation. Since in CLIPstyler an image/image patch is represented using the class token, we analogously use the class token extracted from $\hat{V}$ (i.e., $\pmb{z}_v = \hat{\pmb{v}}_{0}$).

\begin{table}[t]
\centering
\vspace{-0.3cm}
\begin{tabular}{c|c|c}
\toprule
band & Frequency index & Period (tokens) \\
\midrule
$b_1$    & 0-1       & 25-$\infty$             \\
$b_2$   & 2-3       & 7-25            \\
$b_3$    & 4-7       & 4-7             \\
$b_4$    & 8-15      & 2-4             \\
$b_5$    & 16-49     & 1-2            \\
\bottomrule
\end{tabular}
\caption{Correspondences between the bands, frequency indexes and period (tokens). The corresponding periods are approximate numbers of tokens that are needed to complete a cosine wave cycle.}
\label{tab:r1}
\vspace{-0.3cm}
\end{table}

\begin{figure*}[t]
  \centering
  \vspace{-0.2cm}
   \includegraphics[width=\linewidth]{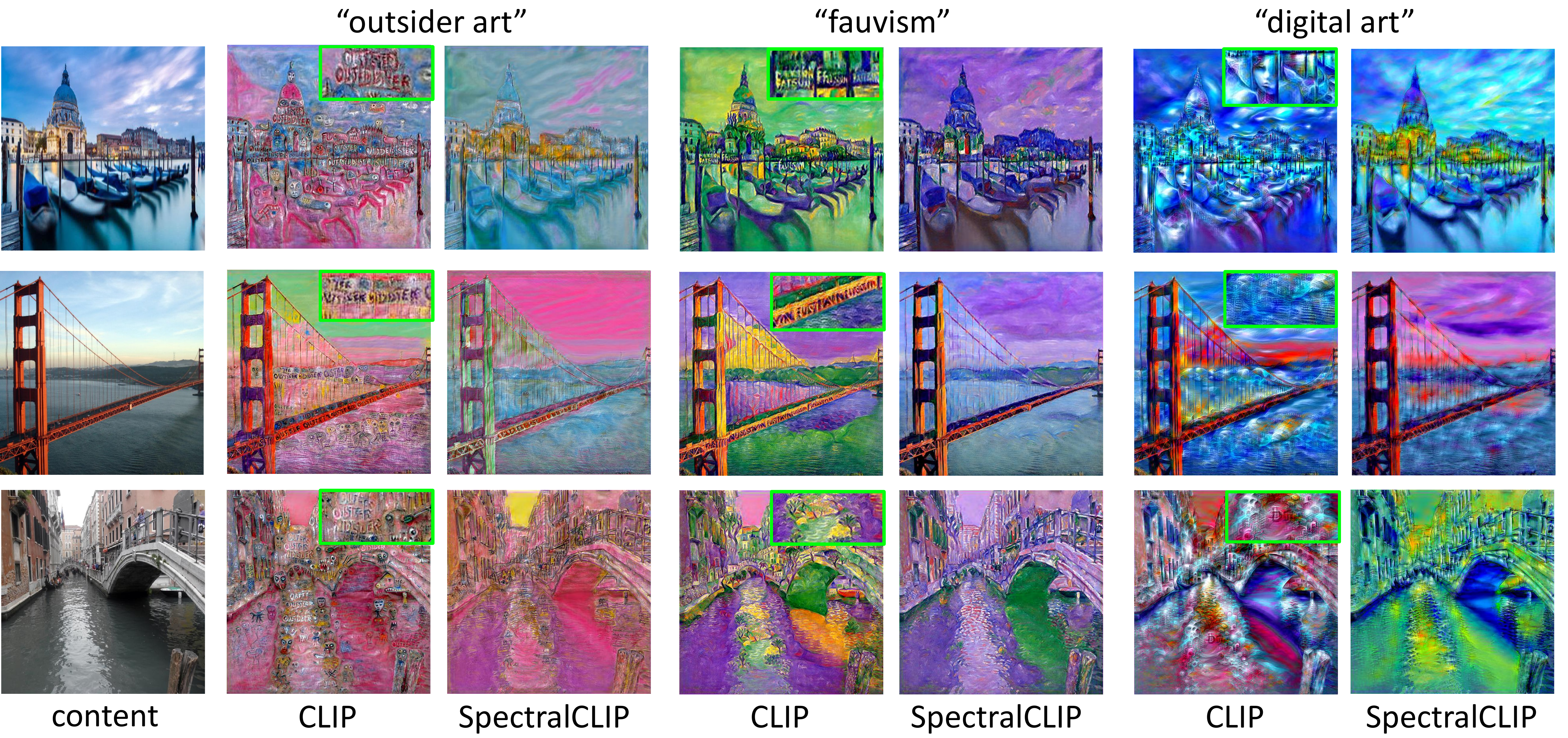}
   \caption{Style transfer results using either the CLIP or  SpectralCLIP to condition the generation. SpectralCLIP effectively prevents artifact generation while achieving similar style features, e.g., color, paint stroke (artifacts are \textit{\textcolor{green}{highlighted}}). }
   \label{fig:exp1}
   \vspace{-0.3cm}
\end{figure*}

\subsection{Band selection}
\label{sec:band-selection}
So far we have assumed that we can associate a frequency filter $\pmb{b}$ to a given textual description of a style (in CLIPstyler) or to a textual prompt (in VQGAN+CLIP). However, since selecting the best $\pmb{b} \in \{0,1\}^n$ would be intractable, we adopt a simpler solution, inspired by \cite{Prism}, where the whole spectrum of frequencies ($0, ..., n-1$) is split in fixed bands, each being a contiguous interval of frequencies. 
Note that each index $m$ of the DCT has a frequency of $2m$ (see \cref{eq:dct}), and that given a sequence of length $N$, the period is $N/2m$ (it takes $N/2m$ tokens to complete a cycle).
For example, in our case where ViT-B/32 is used as the vision encoder, the sequence length is $N = (224/32)^2+1=50$. Therefore, index 1 of the DCT corresponds to the period of 25, and index 5 to the period of 5, etc. 
Specifically, we define the following 5 bands: 
$b_1 = [0,1]$, 
$b_2 = [2,3]$, 
$b_3 = [4,7]$, 
$b_4 = [8,15]$, 
$b_5 = [16,49]$.
The correspondence between these frequency bands and periods is presented in \cref{tab:r1}. 
In ViT-B/32, the input image is first resized to the resolution of $224^2$, and then divided into $7\times7$ image patches of $32^2$. 
In this sense, it can be estimated that $b_1$ relates to artifacts that roughly span more than 3 lines of patches, $b_2$ to those spanning 2 lines, and $b_3-b_5$ to small artifacts spanning within 1 line. 

Another problem is that, given a style description, the artifact condition is unpredictable.
Specifically, it is difficult to judge if the stylized image contains artifacts or not, as well as to detect the artifact appearance.
Nevertheless, through experiments on various styles, we find the artifacts are usually at three scales.
Therefore, we propose a simple yet effective method based on empirical studies.
Through experimenting on multiple band combinations, we find three filtering strategies ($c_1 = \{ b_1, b_2, b_4 \}$,
$c_2 = \{ b_1, b_2 \}$,
$c_3 = \{ b_1\}$) that are effective for preventing the artifacts at the 
corresponding three scales, respectively.
For instance, using $c_1$, the associated filter $\pmb{b}$ contains ones in the intervals $b_1, b_2, b_4$, and it is used in \cref{eq:filtering} to zero out the corresponding bands.
We use visual inspection to select the band combination that leads to the best result, then it
is used in all image stylisation conditioned on $s$.
This selection step is done {\em only once per given style s} using a single image content.
More details are provided in Appendix \ref{band_selection}.



\section{Experiments}
\label{sec:Experiments}

\begin{figure*}[t]
  \centering
  \vspace{-0.2cm}
   \includegraphics[width=\linewidth]{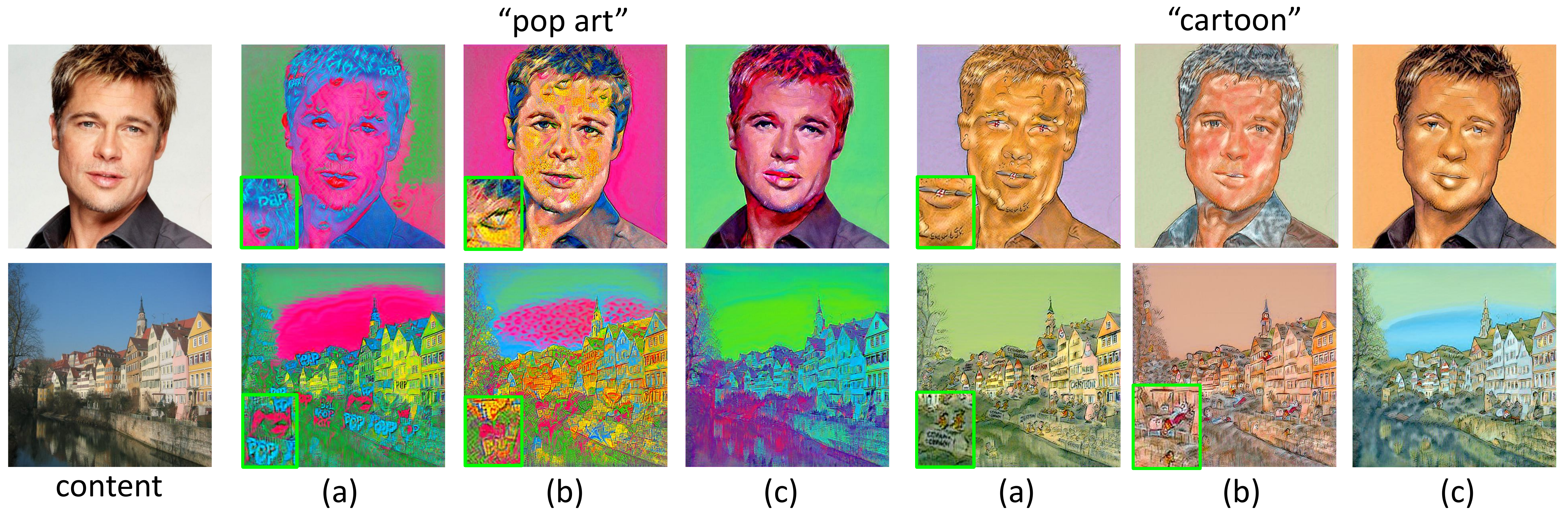}
   \caption{Comparisons among text-guided style transfer results, generated using CLIPstyler with (a) CLIP,   (b) forget-to-spell CLIP \cite{Torralba-Disentangling}, and (c) SpectralCLIP, respectively (artifacts are \textit{\textcolor{green}{highlighted}}).}
   \label{fig:exp2}
  \vspace{-0.4cm}
\end{figure*}

\subsection{Style Transfer Results}
\label{sec:StyleTransferResults}
In this section, we evaluate  SpectralCLIP in a text-guided style transfer task.
For a fair comparison with CLIPstyler\cite{kwon2022clipstyler},
we use its same network, loss functions, training protocols, hyperparameters, etc., changing {\em only}  
 the basic image-text similarity as described in \cref{sec:metrics}.

\noindent\textbf{Qualitative results.}
In \cref{fig:teaser} and \cref{fig:exp1}, we show multiple text-guided style transfer results generated using either the standard CLIP space (i.e., the original CLIPstyler method) or our SpectralCLIP.
These images show that  CLIPstyler frequently generates visual  and textual artifacts.
By contrast, the results generated using SpectralCLIP do not have the issues of both visual artifacts and textual artifacts while the styles are presented well.
Take our results of ``outsider art" (\cref{fig:exp1}) for example, the style shown in the images is similar to the style in the results of CLIPstyler, but with artifacts excluded.
More qualitative results are provided in Appendix \ref{addtional_results}. 
Additionally, results of non-artistic concrete styles (\eg, fire) are also provided in Appendix \ref{concrete_styles}.

\noindent\textbf{Quantitative results.}
Quantitatively evaluating a style-transfer approach is difficult because of the lack of a universally accepted metric that can assess the reflection of a target style in a generated image.
On the other hand, using a cosine similarity between the CLIP-based representations of the target style and the generated image may favour those methods (such as CLIPstyler) which use the same similarity in the optimization stage. To partially solve this problem, we use an additional metric, based on the ``learn-to-spell'' projection of the CLIP space proposed (and trained) in \cite{Torralba-Disentangling} (\cref{sec:Related}). The idea behind this metric is that since part of the artifacts have a textual nature (i.e., strings drawn on the generated images, \cref{sec:intro,sec:Related}), then a learn-to-spell based similarity between a generated image and the corresponding textual description of the style
should be {\em higher} for those images containing {\em more} textual artifacts. We provide more discussion of this metric in Appendix \ref{learn2spell_sim}.

Concretely,  we  sample 100 images from the COCO~\cite{lin2014microsoft} val-set  and use them as content images.
Then, for each style, we generate the style transfer results using either SpectralCLIP or CLIPstyler.
Finally, we use both the original  CLIP space  and learn-to-spell CLIP projection \cite{Torralba-Disentangling} to evaluate the similarity between the textual style description and the generated images.
 \cref{tab:quant} shows that  using SpectralCLIP, the learn-to-spell CLIP score is significantly reduced,
 indicating that SpectralCLIP effectively prevents textual artifact generation.
 On the other hand, the CLIP similarity is also reduced; however, as aforementioned, this metric is biased towards CLIPstyler, where the whole, non-filtered CLIP image representation is used for optimization.

\begin{table}[t]
\begin{tabular}{c|l|l|l}
\toprule
                         &        & \multicolumn{1}{c|}{\begin{tabular}[c]{@{}c@{}}CLIPstyler \\ w. CLIP\end{tabular}} & \multicolumn{1}{c}{\begin{tabular}[c]{@{}c@{}}CLIPstyler w.\\ SpectralCLIP\end{tabular}} \\ \midrule
\multirow{2}{*}{cartoon} & \textit{CLIP}  & {$0.269$ \scriptsize{$\pm 0.014$}}                & {$0.256 $ \scriptsize{$\pm 0.014$}}          \\
                         & \textit{CLIP-Spell} & {$0.482 $ \scriptsize{$\pm 0.056$}}                 & {$0.441 $ \scriptsize{$\pm 0.066$}}          \\ \hline
\multirow{2}{*}{pop art} & \textit{CLIP}   & {$0.315 $ \scriptsize{$\pm 0.023$}}                 & {$0.287 $ \scriptsize{$\pm 0.018$}}          \\
                         & \textit{CLIP-Spell} & {$0.419 $ \scriptsize{$\pm 0.137$}}                 & {$0.353 $ \scriptsize{$\pm 0.121$}}          \\ \hline
\multirow{2}{*}{\begin{tabular}[c]{@{}c@{}}visionary\\ art\end{tabular}} & \textit{CLIP} & {$0.322 $ \scriptsize{$\pm 0.016$}} & {$0.278 $ \scriptsize{$\pm 0.018$}} \\
                         & \textit{CLIP-Spell} & {$0.527 $ \scriptsize{$\pm 0.086$}}                 & {$0.397 $ \scriptsize{$\pm 0.057$}}          \\ \hline
\multirow{2}{*}{\begin{tabular}[c]{@{}c@{}}outsider\\ art\end{tabular}}  & \textit{CLIP} & {$0.314 $ \scriptsize{$\pm 0.018$}} & {$0.255 $ \scriptsize{$\pm 0.019$}} \\
                         & \textit{CLIP-Spell} & {$0.571 $ \scriptsize{$\pm 0.108$}}                 & {$0.360 $ \scriptsize{$\pm 0.095$}}          \\ \bottomrule
\end{tabular}

\caption{Average cosine similarity between the stylized images and the textual description of the style, measured both on the original  CLIP space $(\uparrow)$ and on the learn-to-spell CLIP (in short referred to as CLIP-Spell) $(\downarrow)$.}
\label{tab:quant}
\vspace{-0.3cm}
\end{table}

\subsection{Comparison  with Forget-to-Spell CLIP}
\label{sec:forget-to-spell}

In this section, we compare  SpectralCLIP 
with  ``forget-to-spell'' CLIP~\cite{Torralba-Disentangling},  which is a learned subspace of CLIP semantic space that alleviates the text-image entanglement problem (\cref{sec:intro}). Specifically, we again use CLIPstyler as the baseline and we replace its (standard) CLIP space with 
 the forget-to-spell projection proposed and trained in \cite{Torralba-Disentangling}.
Hence, we compare three methods: (a) CLIPstyler with CLIP (i.e., the original CLIPstyler), (b) CLIPstyler with the forget-to-spell CLIP, and (c) CLIPstyler with SpectralCLIP.

\noindent\textbf{Qualitative results.}
From the qualitative results shown in 
 \cref{fig:exp2}, we draw three conclusions: 1) the images generated by the original CLIPstyler contain both visual  and textual artifacts; 2) using the forget-to-spell CLIP alleviates the  textual artifact issue, but it still generates visual artifacts, which makes the results unlike human created artworks; and 3) similar to the analysis in \cref{sec:StyleTransferResults}, using SpectralCLIP, no  visual nor textual artifacts have been generated, improving the overall quality of the results.

\begin{table}[]
\begin{tabular}{c|c|c|c}
\toprule
         $(\%)$           &  \begin{tabular}[c]{@{}c@{}}w.\\ CLIP\end{tabular}  & \begin{tabular}[c]{@{}c@{}}w.\\ forget-to-spell\end{tabular}  & \begin{tabular}[c]{@{}c@{}} w.\\ SpectralCLIP\end{tabular} \\ \midrule
Overall   &   $34.44$   &      $10.28$                &       $\mathbf{55.28}$            \\ \hline
Artifact-Free &$19.45$    &  $6.11$                     &    $\mathbf{74.44}$     \\ \bottomrule
\end{tabular}
\caption{User preference of the three style transfer methods with respect to the overall quality of the generated images $(\uparrow)$ and the presence of artifacts $(\uparrow)$.}
\label{tab:user_study}
\vspace{-0.4cm}
\end{table}

\begin{figure*}[ht]
  \centering
   \includegraphics[width=\linewidth]{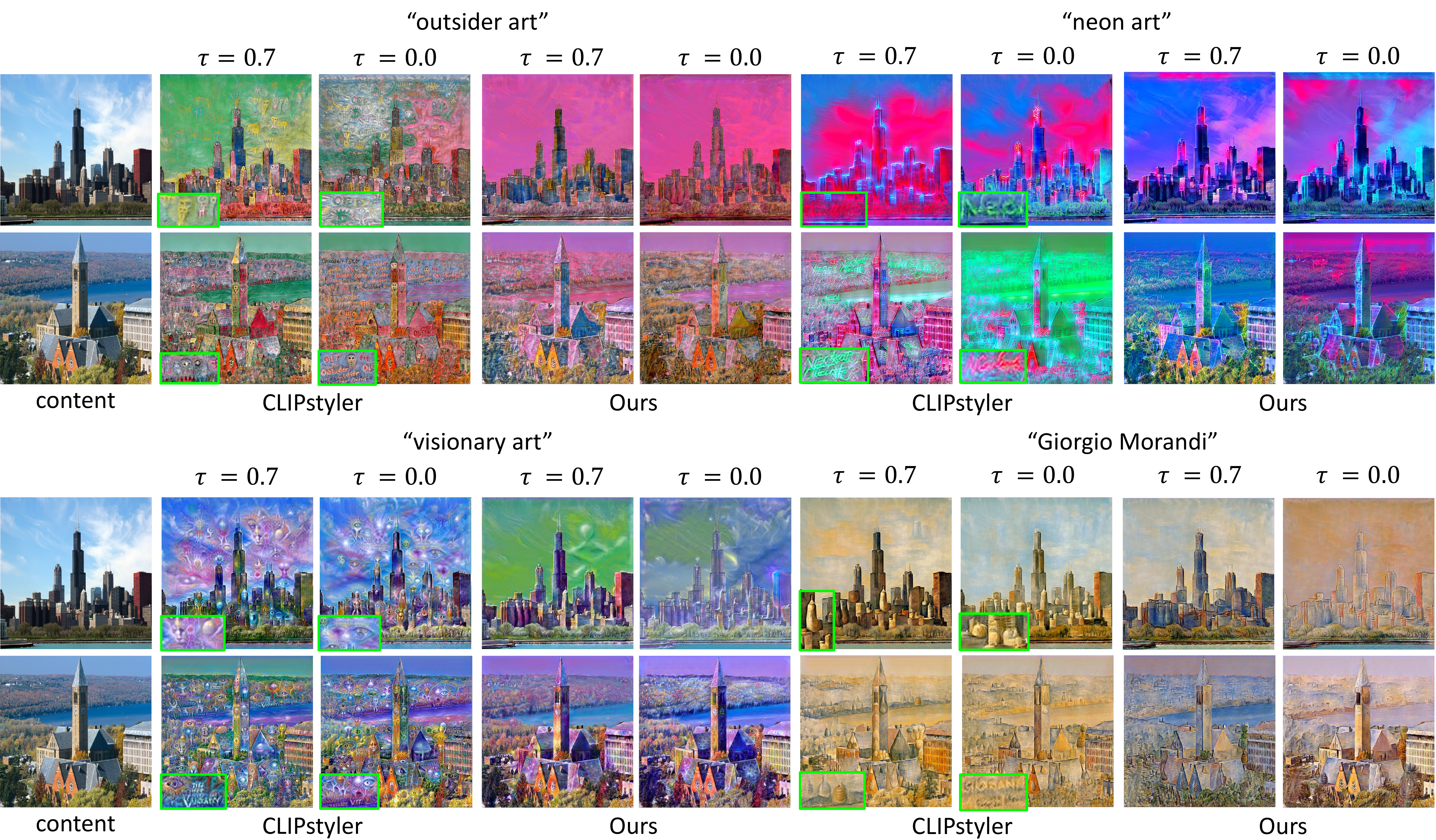}
   \caption{The effect of the patch-rejection threshold $\tau$ in the CLIPstyler patch-loss (artifacts are \textit{\textcolor{green}{highlighted}}, zoom in to see details).}
   \label{fig:exp3}
  \vspace{-0.4cm}
\end{figure*}

\noindent\textbf{User Study.}
We further compare the three methods through a user study.
Specifically, we use 10 styles jointly with two different tasks which respectively analyse: 1) the overall quality of the generated images, asking the users to assess  whether the stylized results are consistent with the target style, the content is well preserved, and no inharmonious artifacts are generated;
and 2) the specific artifact issue, asking the users to assess  the generated images regarding the possible presence of visual/textual artifacts.
We randomly sampled 100 content images from the COCO val-set and then created a questionnaire with 24 questions.
We recruited  30 users, who were asked to
select one out of the three methods for each content image.
The user preference results, as reported in \cref{tab:user_study},
show that  SpectralCLIP gets the best scores on both tasks.
Specifically, SpectralCLIP achieves a significantly higher preference score ($77.44\%$) in the artifact-free evaluation, indicating the effectiveness of our proposal in preventing artifacts.
More details about the user study are provided in Appendix \ref{user_study}.

\subsection{Hyperparameter Study}
\label{hyperparam}
\noindent\textbf{Threshold in the patch loss.}
CLIPstyler uses a patch-rejection threshold $\tau$ in its  patch loss to alleviate  the over-specification problem (\cref{sec:Related}).
The value of this threshold is a (manually selected) hyperparameter, which is fixed to $\tau=0.7$
in \cite{kwon2022clipstyler}.
In \cref{fig:exp3} we compare the use of this threshold ($\tau=0.7$) with a non-thresholding variant  ($\tau=0.0$).
The results show that the original CLIPstyler is heavily influenced by the thresholding,
since its removal ($\tau=0.0$) leads to the generation of many  more artifacts, independently of the target style.
By contrast,  SpectralCLIP is much less sensitive to this thresholding, since
 the results using $\tau=0.0$ are consistent with the images generated with  $\tau=0.7$, showing that our method does not rely on this thresholding step to avoid over-specification.

\begin{figure}[t]
  \centering
  \vspace{-0.2cm}
   \includegraphics[width=.9\linewidth]{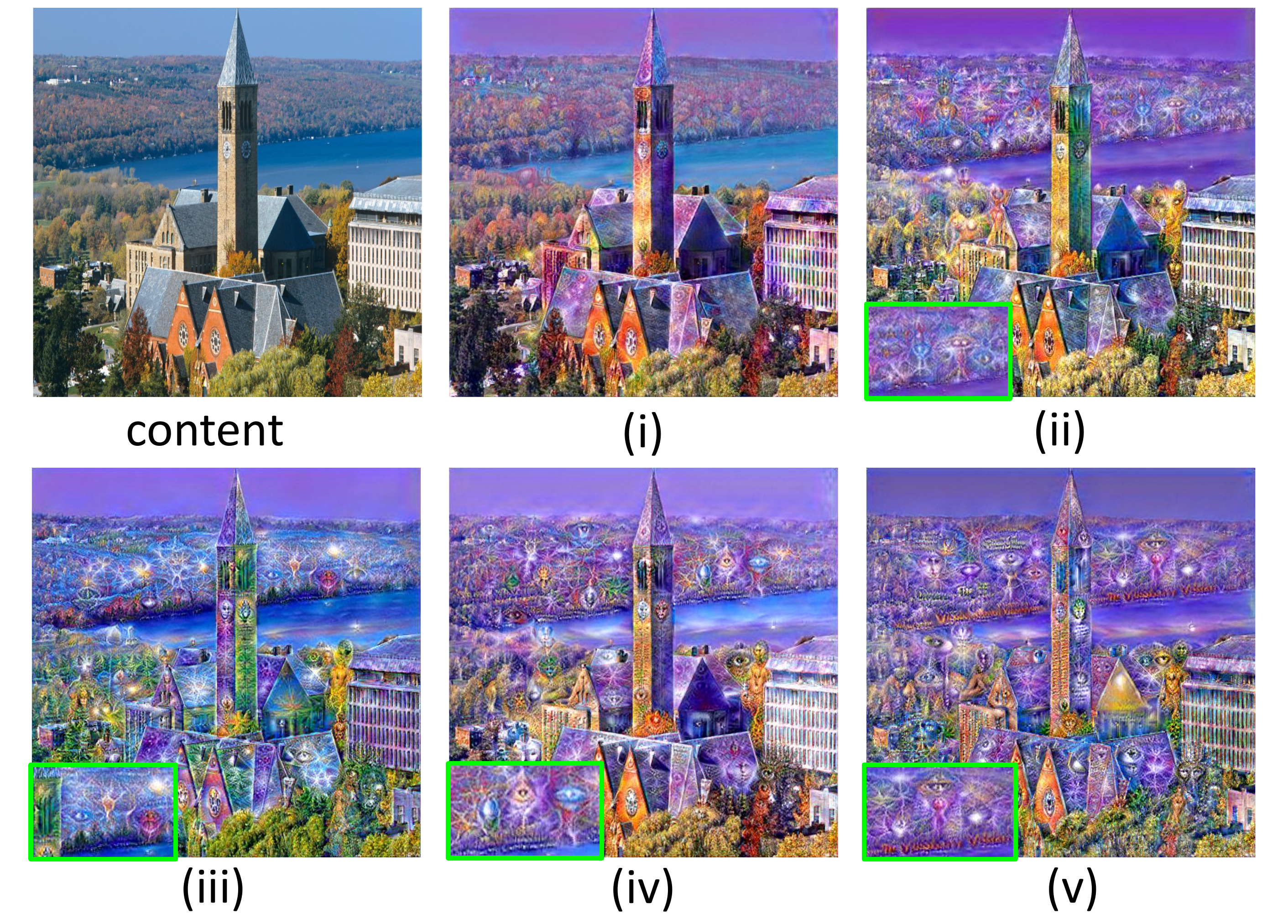}
   \caption{The effects of different band-stop filters (artifacts are \textit{\textcolor{green}{highlighted}}).}
   \label{fig:exp4}
  \vspace{-0.3cm}
\end{figure}

\noindent\textbf{Band selection.}
We  study the effect of the band selection (\cref{sec:band-selection}) using 
the style ``visionary art" (\cref{fig:exp4}), jointly with  five different filters: (\romannumeral1) masking bands 1, 2 and 4 (corresponding to $c_1$ in \cref{sec:band-selection});
(\romannumeral2) masking bands 1, 2 and 5;
(\romannumeral3) masking bands 1 and 2 ($c_2$ in \cref{sec:band-selection}); 
(\romannumeral4) masking bands 1 ($c_3$ in \cref{sec:band-selection});
and (\romannumeral5) only masking the lowest frequency (i.e., the frequency index $m=0$).
 \cref{fig:exp4} shows that 
 the  visual appearance change caused by higher frequencies tends to be more local, as can be observed by comparing  (i) with (ii), and  (ii) with (iii).
For instance,
comparing
(i) with (ii), the former filter leads to greater visual appearance changes within a larger region.
Moreover, the differences between (ii) and (iii) are marginal, indicating that  band 5 (which includes the highest frequencies) is related to changes in smaller areas.
Furthermore, lower frequencies result in the generation of larger artifacts, as shown by the comparison  between (iii) and (iv), and between  (iv) and (v).
For example,
not masking band 2 leads to the generation of larger and more obvious artifacts on the eyes.
A similar phenomenon can be observed when we additionally do  not mask the frequency with index 1 (see (iv) and (v)).
These  results confirm  our assumption (\cref{sec:intro}) that most artifacts are visual structures with a  specific period, whose generation  can be prevented by adopting the proposed spectral representation and the corresponding frequency filters.

\subsection{Text-to-Image Generation}
\label{image_generation}
\begin{figure}[t]
  \centering
   \includegraphics[width=.75\linewidth]{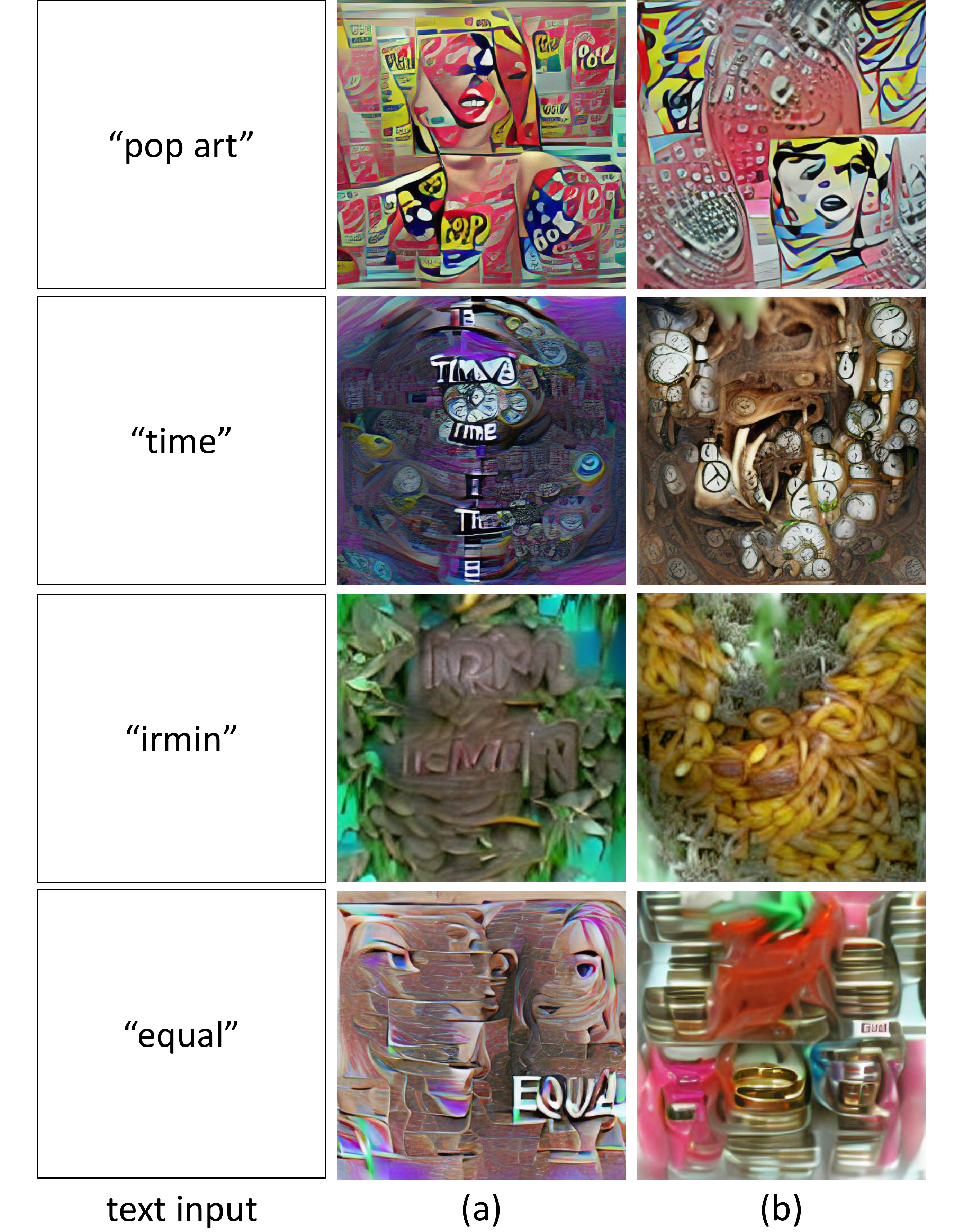}
   \caption{Text-to-image generation results using (a) VQGAN+CLIP, and (b) VQGAN+SpectralCLIP.}
   \label{fig:exp5}
   \vspace{-0.4cm}
\end{figure}

To test the generality of SpectralCLIP, we consider  
a different task, adopting the text-to-image generation framework used in \cite{Torralba-Disentangling} to evaluate the word-image disentanglement of forget-to-spell and learn-to-spell (\cref{sec:intro,sec:forget-to-spell}).
Specifically, following \cite{Torralba-Disentangling}, we use  VQGAN+CLIP~\cite{crowson2022vqgan}
and we replace its text-image similarity computed on the original CLIP space
with  SpectralCLIP.
\cref{fig:exp5} compares the results obtained with VQGAN+CLIP and VQGAN+SpectralCLIP, 
and confirms the observations of Materzynska \etal~\cite{Torralba-Disentangling}, who highlight that 
VQGAN+CLIP frequently generates inappropriate textual strings (textual artifacts) mixed with visual content.
By contrast, this problem is largely alleviated with  SpectralCLIP.
Meanwhile, the generated image content in VQGAN+SpectralCLIP is still consistent with the given text prompt (except for the nonsense text input ``irmin").
In all the VQGAN+SpectralCLIP results shown in \cref{fig:exp5},
we use the same filtering strategy (masking only  band 4, i.e., $b_4$, see \cref{sec:band-selection}). Note that  the scale of a textual/visual artifact depends on the CLIP encoder input, which is the full image in the case of VQGAN,  and  this results in a shorter period with respect to CLIPstyler.

\paragraph{Conclusion.}
Despite the wide success of vision-language foundation models like CLIP in different vision-language tasks, directly using CLIP for style transfer suffers from the generation of visual and textual  artifacts.
To resolve this problem,  we propose SpectralCLIP, which transforms the CLIP embedding sequence into the frequency domain and filters those frequencies whose period corresponds to the artifact scales.  
Experimental style transfer results show that SpectralCLIP significantly mitigates artifact generation, thus improving the realistic degree and the quality of the generated images. 

\paragraph{Limitations.}
Despite the promising results of SpectralCLIP, there are still some limitations.
Firstly, we empirically analyse the artifact patterns present in a range of artistic styles, and mask out certain bands using one of the three general filters. 
The reason why a certain target style tends to produce different scales of artifacts is still unclear. This may require a deeper understanding of how CLIP captures these artistic concepts when it was pre-trained. 
Secondly, this work defines three general band combinations that effectively produce cleaner stylised images. 
A more promising alternative for future work is to automatically select frequency bands that cater to a target style.
Recently, in the language domain, M{\"u}ller-Eberstein \etal \cite{muller-eberstein-etal-2022-spectral} promote \cite{Prism} and develop learnable filters rather than handcrafted ones,
offering an intriguing direction to follow. 
To this end, for image style transfer, a widely recognised metric to measure the presence of artifacts is still missing.

\paragraph{Acknowledgment.}
This work was supported by the MUR PNRR project FAIR (PE00000013) funded by the NextGenerationEU, 
by the PRIN project CREATIVE (Prot. 2020ZSL9F9), and by the EU Horizon project ELIAS (No. 101120237).

{\small
\bibliographystyle{ieee_fullname}
\bibliography{egbib}
}

\clearpage
\appendix
\noindent\textbf{\huge Appendix}
\section{Band Selection Details}
\label{band_selection}
As mentioned in \cref{sec:band-selection}, for each target style we  select one of the three band combinations $c_1, c_2$ or $c_3$, and then we use the corresponding filter for all the images translation of that style. 
Note that this is a negligible operation if compared, \eg, to the collection of a set of reference images used to train a style-specific generator  \cite{sanakoyeu2018styleaware, zhu2017unpaired, kotovenko2019content}.
Moreover, there are two underlying reasons: 1) given a style description, the artifact condition is unpredictable.
To be specific, it is difficult to judge if the stylized image contains artifact or not, as well as to detect the artifact appearance; 2) the artifact scales are of limited variable ranges.

\begin{figure}[h]
\vspace{-0.3cm}
  \centering
  \includegraphics[width=0.65\linewidth]{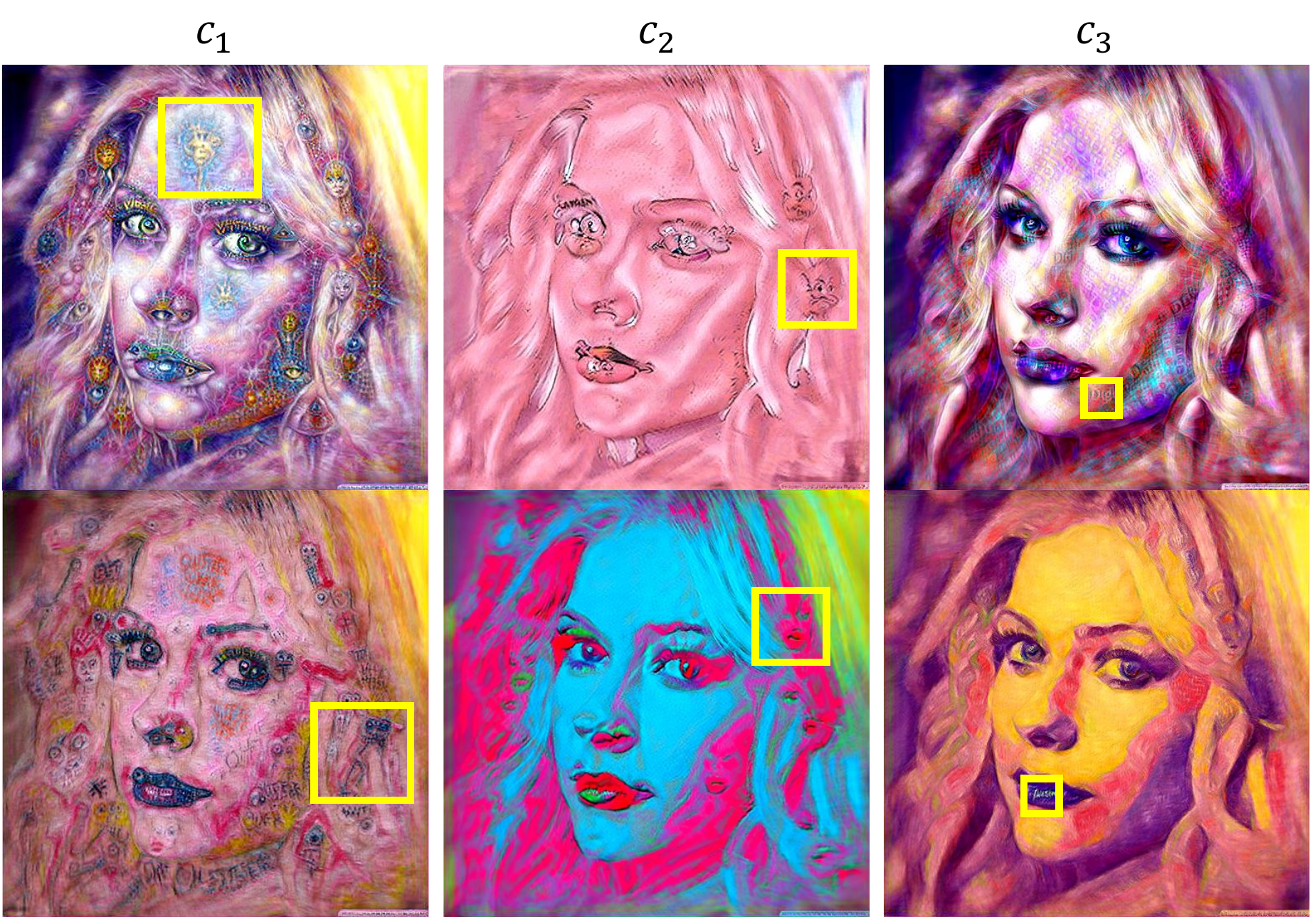}
  \caption{Different styles result in artifacts at different scales which are tricky to predict.}
  \label{fig:r1}
  \vspace{-0.35cm}
\end{figure}

Hence, 
we propose SpectralCLIP, a simple yet effective method based on empirical studies.
Through experimenting on multiple band combinations, we find three filtering strategies ($c_1 = \{ b_1, b_2, b_4 \}$,
$c_2 = \{ b_1, b_2 \}$,
$c_3 = \{ b_1\}$) that are effective for preventing the artifacts at the 
corresponding three scales (as shown in Fig.~\ref{fig:r1}) (and 
through experiments on various styles, we find the artifacts are usually of one of the three scales).
To prevent artifacts of larger scales, more frequency bands should be masked.
The bands are defined according to the scales of the artifacts, i.e., the length of the dependant visual tokens in the CLIP-ViT scenario.

\begin{figure*}[ht]
  \centering
  \begin{subfigure}[]{\linewidth}
    \includegraphics[width=\linewidth]{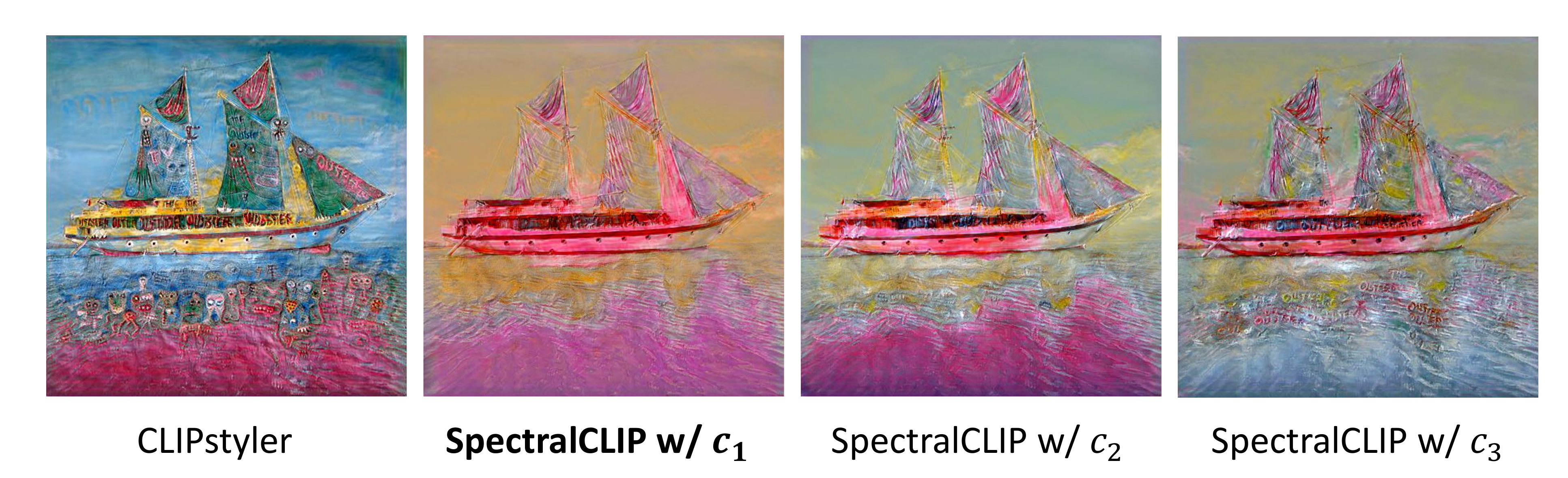}
    \caption{``outsider art"}
    \label{fig:a0_out}
\end{subfigure}

\begin{subfigure}[]{\linewidth}
    \includegraphics[width=\linewidth]{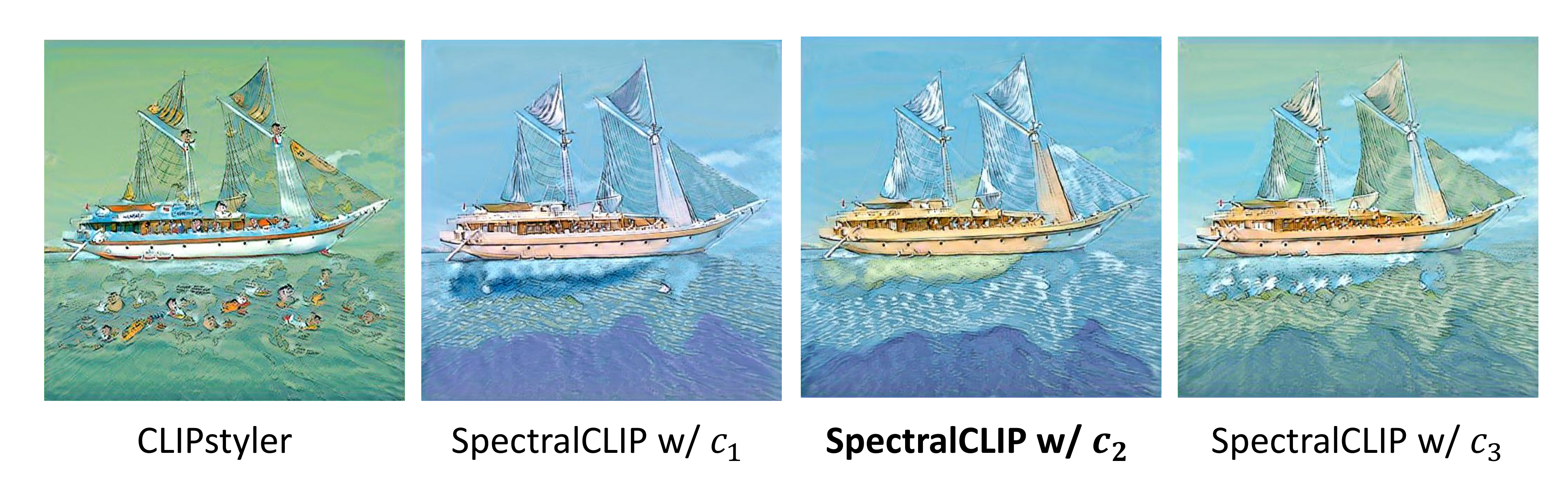}
    \caption{``cartoon"}
    \label{fig:a0_car}
  \end{subfigure}

\begin{subfigure}[]{\linewidth}
    \includegraphics[width=\linewidth]{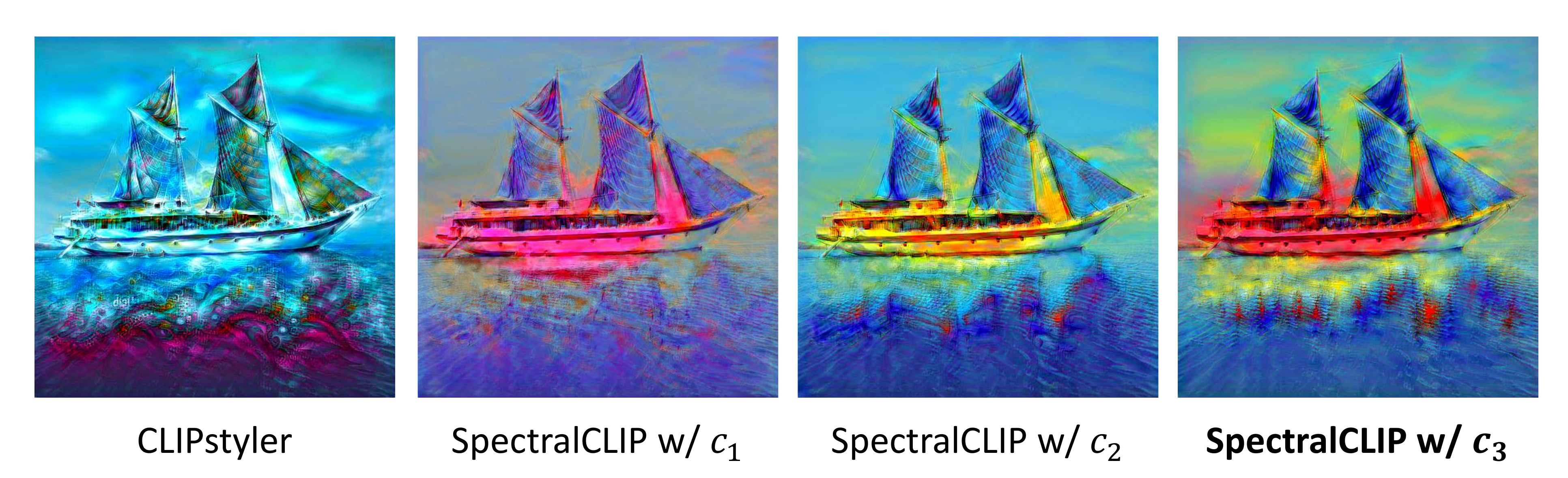}
    \caption{``digital art"}
    \label{fig:a0_dig}
  \end{subfigure}
  
  \caption{Leftmost column: images generated using CLIPstyler and three different styles. All the other columns show the results obtained using SpectralCLIP and one among $c_1, c_2$ or $c_3$ as the filter. In case of ``outsider art", $c_1$ is the best band combination, while  $c_2$  is the best for ``cartoon", and  $c_3$  for ``digital art".} 
  \label{fig:a0}
\end{figure*}

In the leftmost column of \cref{fig:a0}, we show the stylized results of CLIPstyler corresponding to three different style descriptions (``outsider art", ``cartoon" and ``digital art").
In all cases, CLIPstyler generates a lot of artifacts. Additionally, the corresponding scales vary from style to style. A simple comparison can be seen in Fig.~\ref{fig:r1}. As mentioned in \cref{sec:band-selection}, in order to select the most suitable filter for each style, we use a single content image per style and we generate stylized images using our SpectralCLIP and one among $c_1, c_2$ and $c_3$. 
Correspondingly, we give an illustration in \cref{fig:a0},
where we show the results obtained using SpectralCLIP with one filtering band combination among $c_1, c_2$ or $c_3$. For ``outsider art", $c_1$ is the best band combination, while  $c_2$  is the best for ``cartoon", and  $c_3$  for ``digital art".

\begin{figure}[h]
\vspace{-0.4cm}
  \centering
  \includegraphics[width=0.95\linewidth]{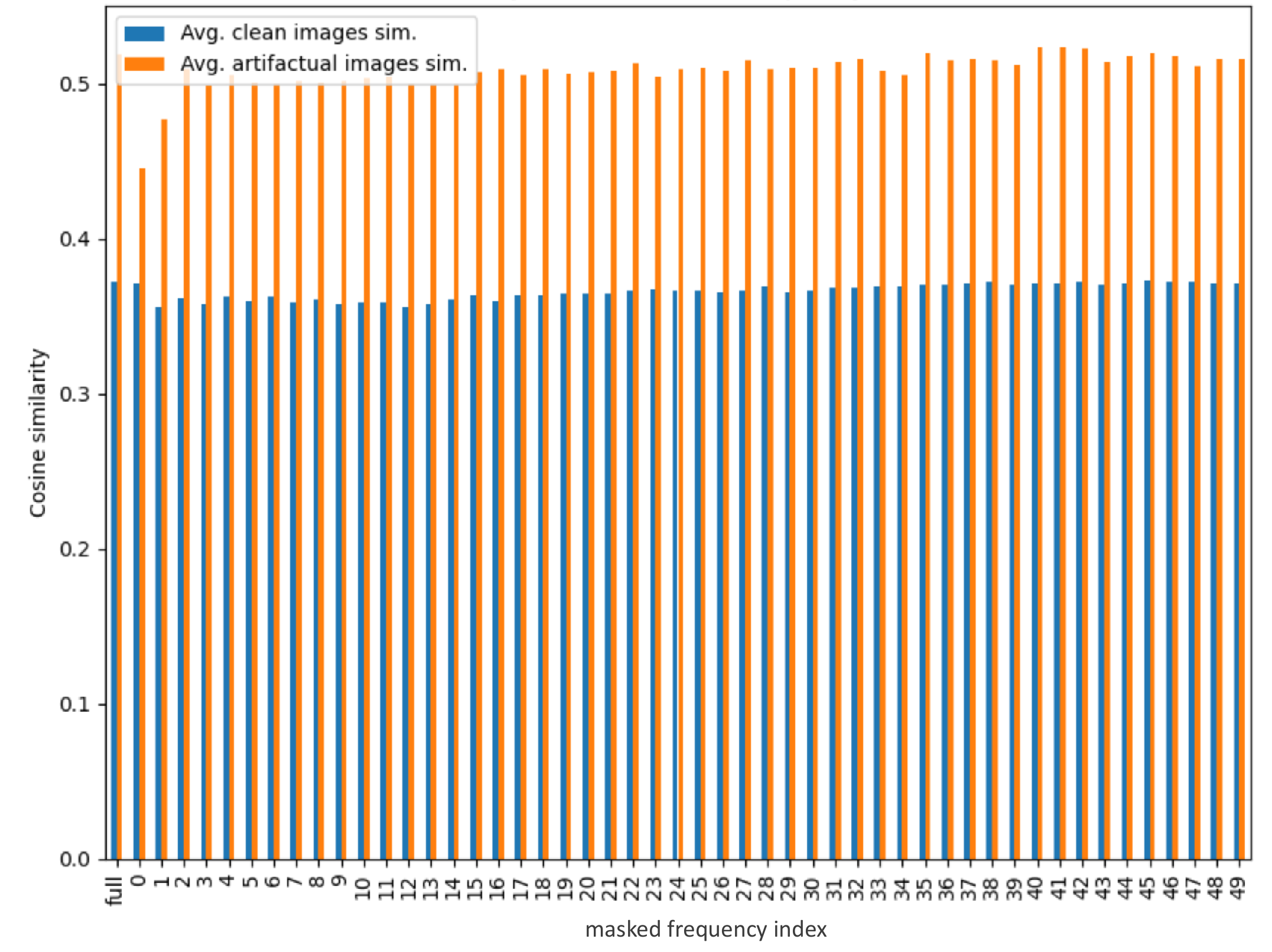}
  \caption{Spectral analysis w.r.t. textual artifacts.}
  \label{fig:r2}
  \vspace{-0.5cm}
\end{figure}

In addition, since we can use the CLIP/forget-to-spell CLIP to generate stylized images with/without text artifacts and use learn-to-spell CLIP to measure the text artifact presence, we did a spectral analysis to find further support for our work.
Specifically, 
We first collect 100 images with/without artifacts using CLIP/forget-to-spell CLIP, respectively.
Then, we individually mask each frequency at one time and use the filtered CLIP representation to compute the cosine similarity using learn-to-spell CLIP (the computations are based on patches as in CLIPstyler).
As shown in Fig.~\ref{fig:r2}, masking frequency 0 and 1 significantly reduces the learn-to-spell CLIP similarity scores of the images containing textual artifacts, indicating that those frequencies are related to text artifacts (e.g., in Fig.~\ref{fig:r1}, masking $c_3$ is useful to prevent the generation of text artifacts).

In \cref{tab:band}, we provide the band combination we used for each of the styles presented in this paper.

\begin{figure}[ht]
  \centering
  \includegraphics[width=.875\linewidth]{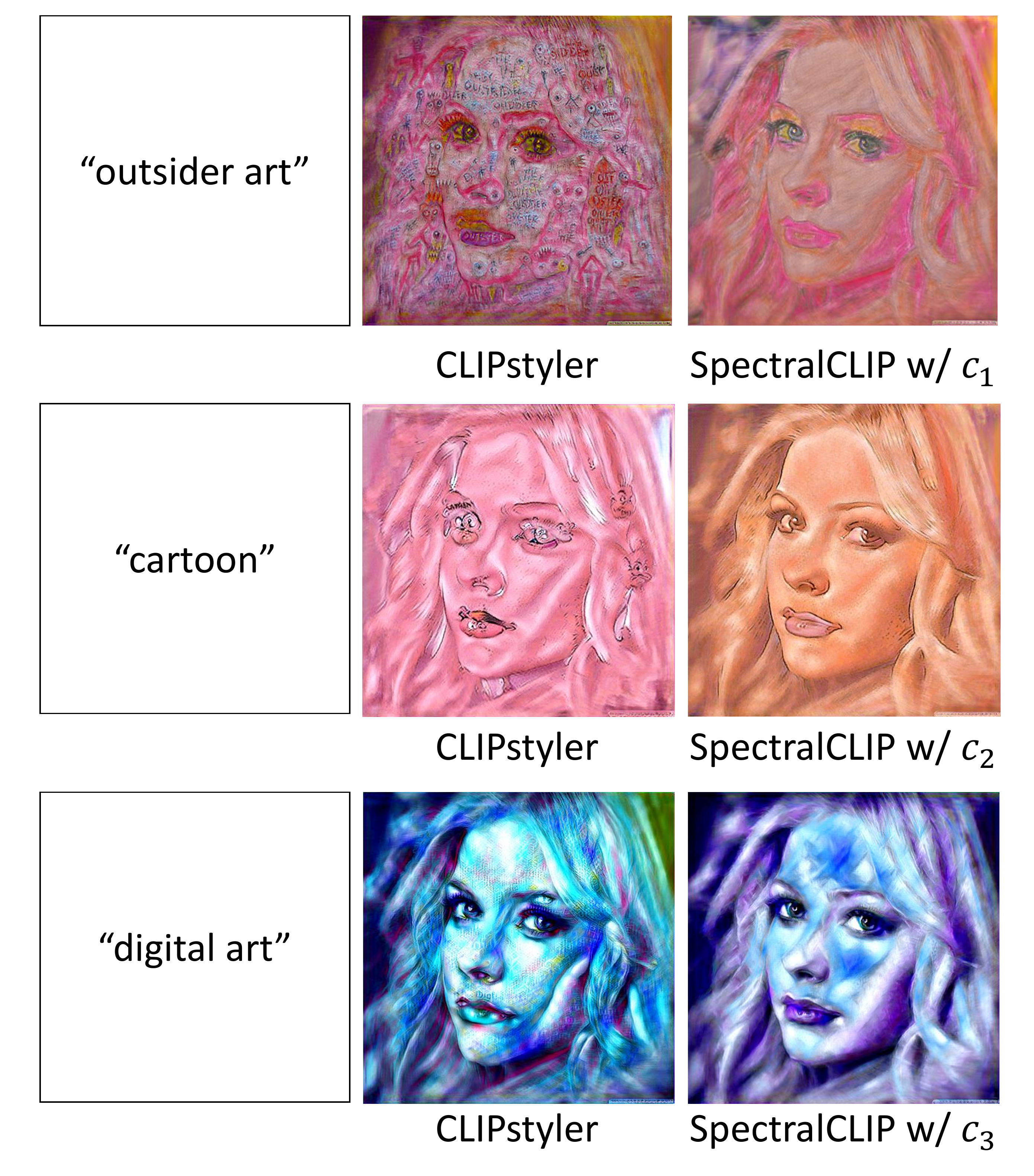}
  \caption{By CLIPstyler, different styles lead to different types of artifacts. Therefore, in SpectralCLIP, we filter different band combinations (indicated on the bottom).}
  \label{fig:aillu0}
\end{figure}

\begin{table}[ht]
\centering
\begin{tabular}{cc}
\toprule
\multicolumn{1}{l|}{Band combination}       & Style                   \\ \midrule
\multicolumn{1}{c|}{\multirow{5}{*}{$c_1$}} & Lowbrow                 \\
\multicolumn{1}{c|}{}                    & Outsider art            \\
\multicolumn{1}{c|}{}                    & Visionary art           \\
\multicolumn{1}{c|}{}                    & Rosy-color oil painting \\
\multicolumn{1}{c|}{}                    & Makoto Shinkai          \\ \hline
\multicolumn{1}{c|}{\multirow{7}{*}{$c_2$}} & Pop art                 \\
\multicolumn{1}{c|}{}                    & Cartoon                 \\
\multicolumn{1}{c|}{}                    & Giorgio Morandi         \\
\multicolumn{1}{c|}{}                    & Harlem renaissance      \\
\multicolumn{1}{c|}{}                    & Neon art                \\
\multicolumn{1}{c|}{}                    & Contemporary art        \\
\multicolumn{1}{c|}{}                    & Francoise Nielly        \\ \hline
\multicolumn{1}{c|}{\multirow{2}{*}{$c_3$}} & Fauvism                 \\
\multicolumn{1}{c|}{}                    & Digital art             \\ \bottomrule
\end{tabular}
\caption{Band combination used for each style.}
\label{tab:band}
\end{table}

\section{User Study Details}
\label{user_study}

\begin{figure*}[t]
  \centering
  \begin{subfigure}[]{.5\linewidth}
    \includegraphics[width=\linewidth]{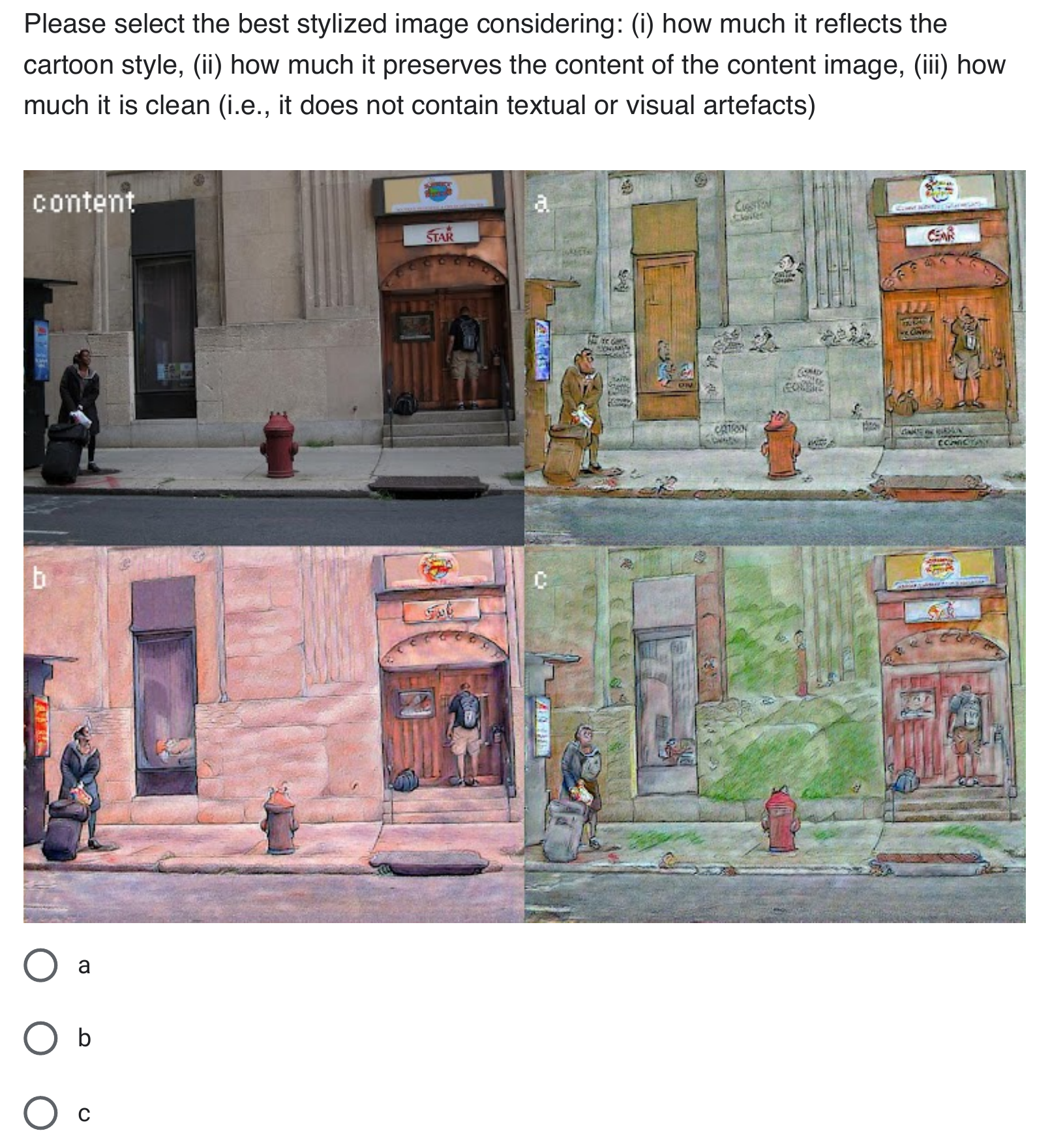}
    \caption{}
    \label{fig:aus1}
\end{subfigure}

\begin{subfigure}[]{.5\linewidth}
    \includegraphics[width=\linewidth]{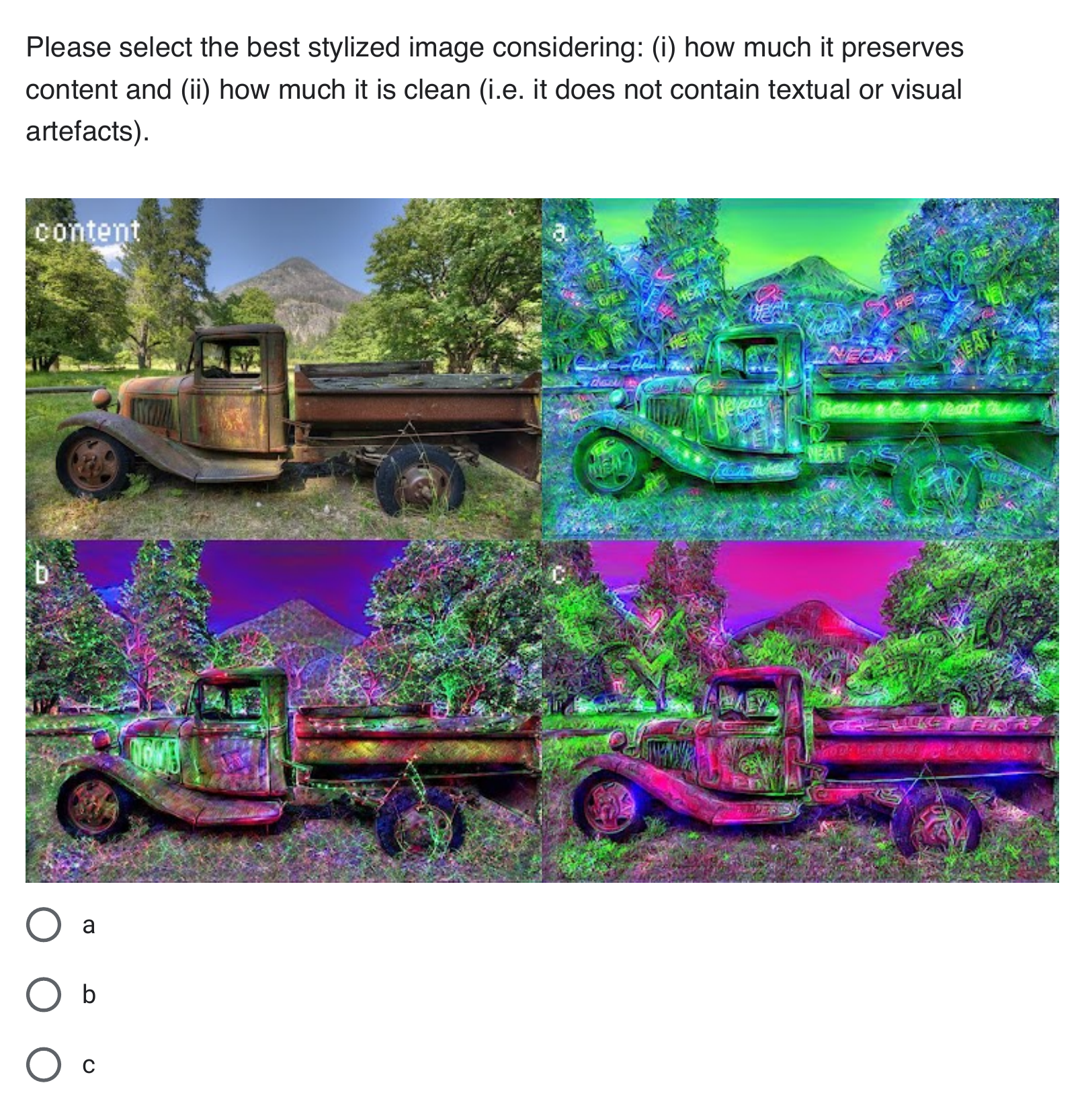}
    \caption{}
    \label{fig:aus2}
  \end{subfigure}
  \caption{The user study questions used to evaluate  two tasks: (a) the overall image quality, and (b) the possible appearance of artifacts.}
  \label{fig:aus}
\end{figure*}



In this section, we provide more details on the user study reported in \cref{sec:forget-to-spell}, in which
we  compare the generation results obtained using text-image similarities computed with three different spaces:  the original CLIP space (which corresponds to the original CLIPStyler), forget-to-spell CLIP, and our SpectralCLIP.

We asked 30 participants to answer 24 questions, split in two tasks (12 questions per task), respectively evaluating the overall quality of the generated images and the possible appearance of artifacts. \cref{fig:aus} shows an example of the both tasks. 
Since evaluating the overall quality of a style-transfer task requires the participants to  consider whether the stylized results reflect target style, for the first task we selected 4 better-known styles (i.e., ``pop art", ``cartoon", ``fauvism" and ``Giorgio Morandi").
For the second task (artifact evaluation), we used the other 6 styles (``lowborw", ``outsider art", ``visionary art", ``harlem renaissance", ``neon art" and ``digital art").
For each style, we randomly sampled 10 images from the COCO val-set and used them as content images to generate the stylized images with the three methods.
Thus, we obtain 100 groups of stylized results in total (each composed of 3 images generated by the 3 compared methods, e.g., see \cref{fig:aus}).
For the first task,  we equally sampled 3 groups of stylzied results for each style, obtaining 12 questions. For the second task, we randomly sampled 12 groups of images without  considering the style.
We used Google Forms as the platform.

\section{Computation Time}
The DCT and IDCT transforms of SpectralCLIP are the only overhead with respect to baselines. 
In text-guided image style transfer, included our method is 1.095 times slower than the CLIPstyler baseline.

\section{Discussion on learn-to-spell Similarity}
\label{learn2spell_sim}
Materzynska \etal~\cite{Torralba-Disentangling} 
analyse the  entanglement problem of written texts and visual concepts in the CLIP space, and learn two
orthogonal projections,
i.e., ``forget-to-spell''  and ``learn-to-spell'',
the former is for recognizing visual concepts while the latter is for recognizing written text. 
Specifically, in this paper, we use the ``learn-to-spell'' projection to measure the textual artifact condition in the generated stylized image, as the more the image contains textual artifacts, the higher the similarity between the image and  the textual description of style  using the ``learn-to-spell'' projection (as in Tab.~\ref{tab:quant} and Fig.~\ref{fig:r2}).

\section{Additional Style Transfer Results}
\label{addtional_results}
\begin{figure*}[!ht]
  \centering
   \includegraphics[width=\linewidth]{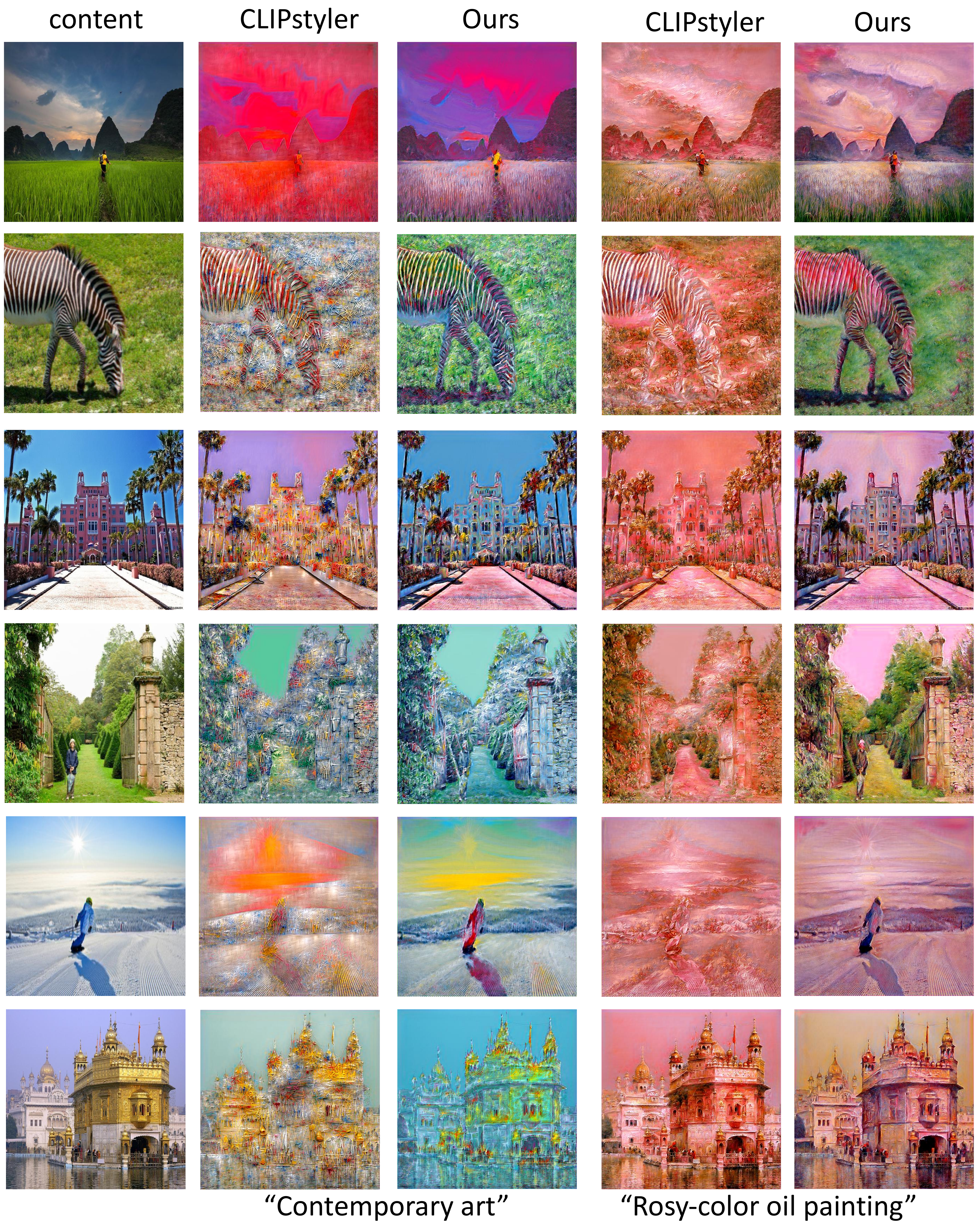}
   \caption{Additional style transfer results using ``Contemporary art" and ``Rosy-color oil painting" as the target textual descriptions.}
   \label{fig:aexp3}
\end{figure*}

\begin{figure*}[!ht]
  \centering
   \includegraphics[width=\linewidth]{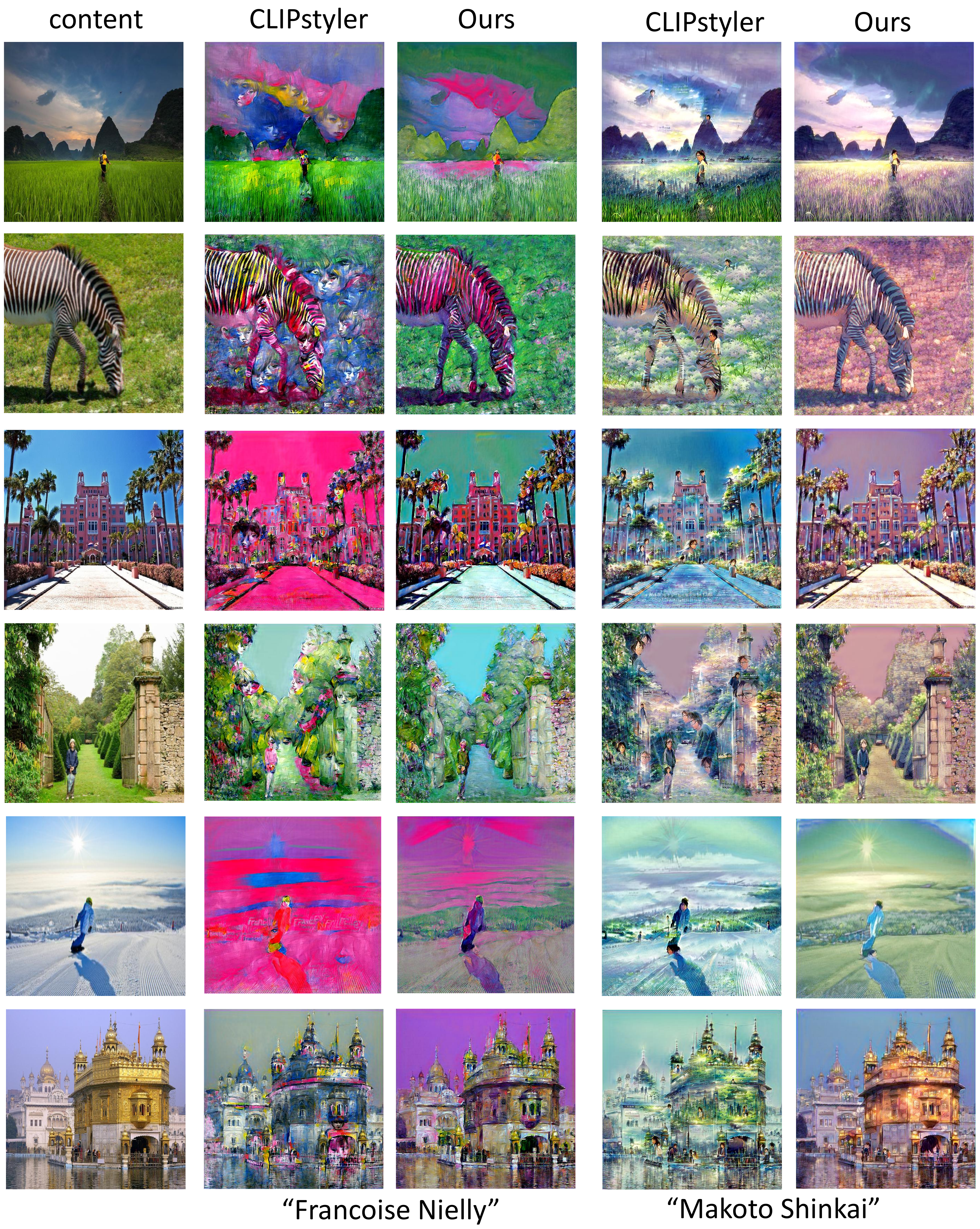}
   \caption{Additional style transfer results using ``Francoise Nielly" and ``Makoto Shinkai" as the target textual descriptions.}
   \label{fig:aexp2}
\end{figure*}

\begin{figure*}[!ht]
  \centering
   \includegraphics[width=\linewidth]{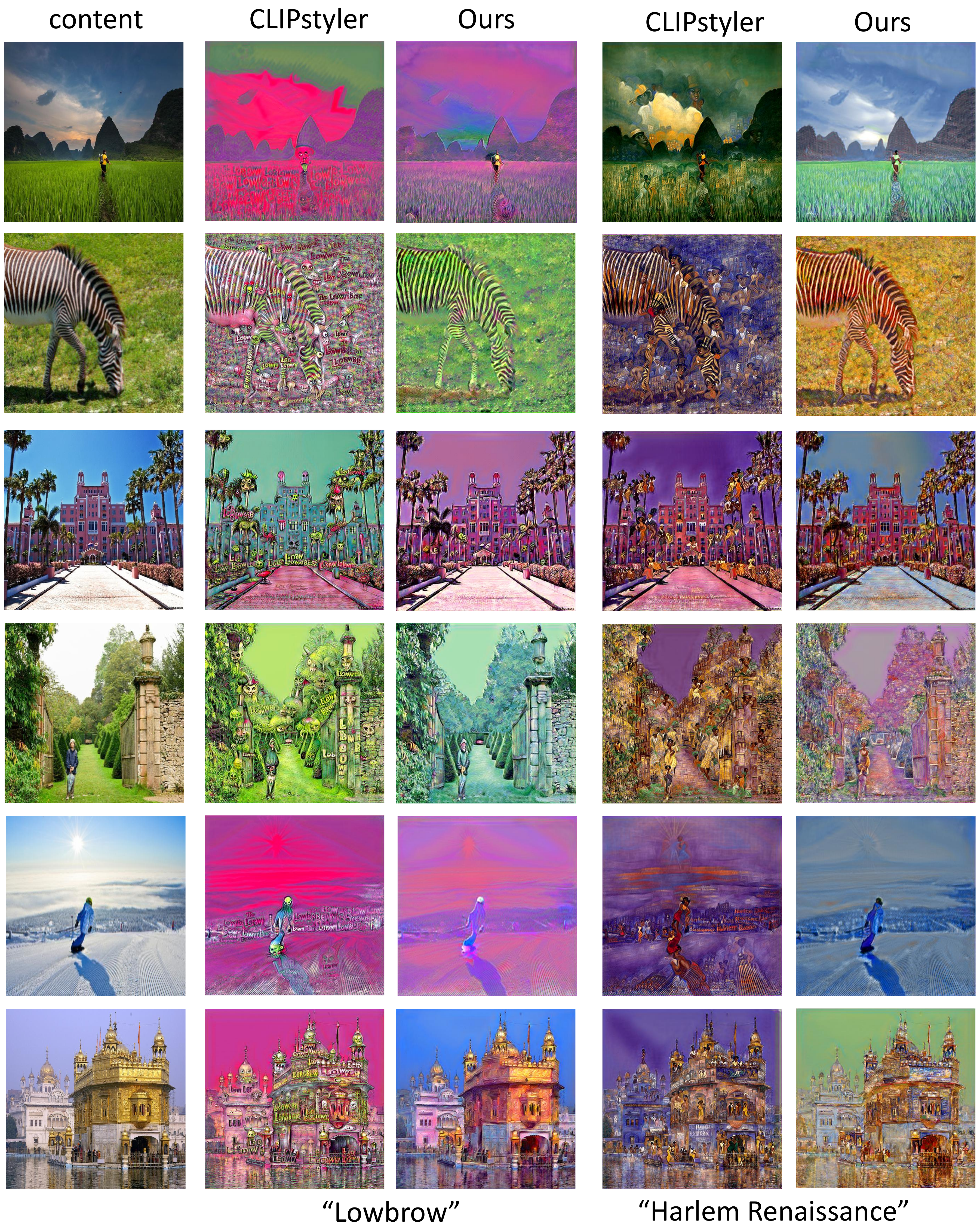}
   \caption{Additional style transfer results using ``Lowbrow" and ``Harlem Renaissance" as the target textual descriptions.}
   \label{fig:aexp1}
\end{figure*}

In this section, we  show additional qualitative  comparisons between the original CLIPstyler and our SpectralCLIP, using as the following target styles: ``contemporary art", ``rosy-color oil painting" (\cref{fig:aexp3}),  ``Francois Nielly", ``Makoto Shikai" (\cref{fig:aexp2}), ``lowbrow", ``harlem renaissance" (\cref{fig:aexp1}).
The results shown in this section confirm those reported in the main paper, and they show that
SpectralCLIP {\em drastically} reduces the 
 generation of both visual and textual artifacts, while simultaneously leading to a high
 consistency of the generated images with respect to the target style. 
For instance, when using CLIPstyler and the ``rosy-color oil painting" 
style (\cref{fig:aexp3})
a lot of 
roses are generated in the background, the sky, the mountains, the trees, etc. 
As another example, in \cref{fig:aexp2}, CLIPstyler ``writes'' the name of the corresponding artist  in the stylized images.
These visual and textual artifacts are definitely not part of the user's desired style, which degrades the quality of the stylised images.
In contrast, the corresponding images  generated  using SpectralCLIP largely solve this problem, making the generation quality significantly higher. 

\section{Non-artistic Concrete Styles}
\label{concrete_styles}
\begin{figure*}[t]
  \centering
  \includegraphics[width=.7\linewidth]{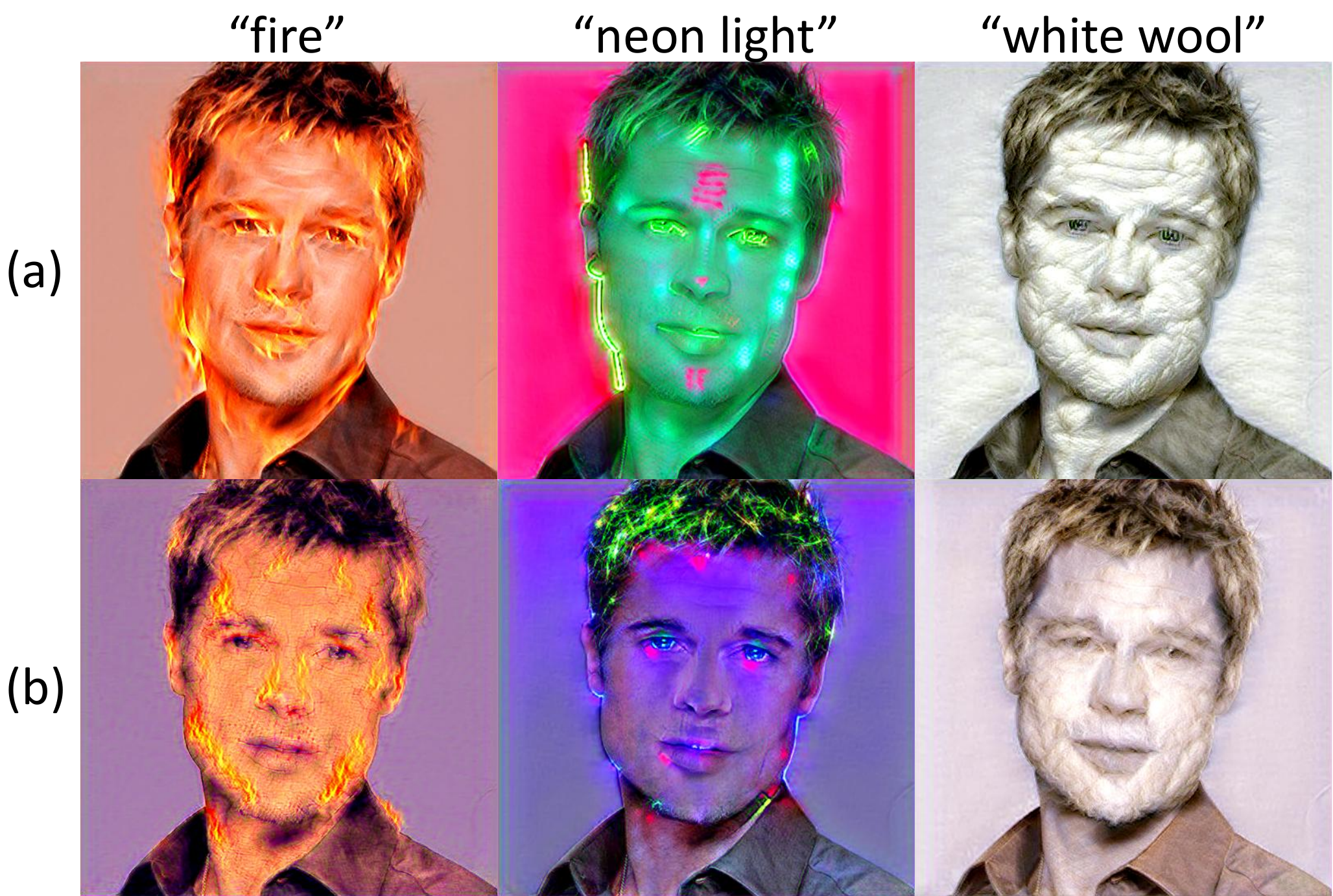}
  \caption{Non-artistic style results of (a) CLIPstyler and (b) SpectralCLIP w. $c_3$.}
  \label{fig:r3}
\end{figure*}

In this work, we focus on artistic and abstract styles, which is a major advantage of CLIP-guided style transfer and yet tends to produce artifacts. 
This section tests the ability of SpectralCLIP to transfer concrete styles, which is also considered in previous style transfer work. We use three concrete styles (``fire'', ``neon light'', and ``white wool'') and report the results in Fig.~\ref{fig:r3}. It can be seen that SpectralCLIP leads to finer-grained stylised results.

\end{document}